\documentclass[11pt]{article}

\usepackage[preprint]{acl}

\usepackage{times}
\usepackage{latexsym}

\usepackage[T1]{fontenc}

\usepackage[utf8]{inputenc}

\usepackage{xcolor}
\usepackage[most]{tcolorbox}
\usepackage{fontawesome5}
\newtcolorbox{profilebox}[2][]{
  colback=white, colframe=gray!20, fonttitle=\bfseries\sffamily,
  coltitle=black, title={#2}, #1
}
\usepackage{enumitem}   

\usepackage{threeparttable}

\definecolor{ai2pink}{HTML}{f0529c}
\definecolor{bg-gray}{gray}{0.95} % 浅灰色背景
\definecolor{prompt}{RGB}{255,221,87} 
\definecolor{diffred}{RGB}{180, 0, 0}
\definecolor{diffgreen}{RGB}{0, 120, 0}
\definecolor{rowgray}{gray}{0.92}
\definecolor{bg-prompt}{HTML}{F0F8FF} 
\definecolor{s1_frame}{RGB}{28, 78, 128}   % 深蓝
\definecolor{s1_back}{RGB}{245, 248, 250}  % 极淡蓝灰
\definecolor{code_var}{RGB}{180, 0, 0}        % 变量颜色 (红)
\definecolor{code_tag}{RGB}{0, 0, 128}  
\definecolor{bg-pink}{HTML}{FFF0F5} % 极淡的粉色背景，比灰色更搭配粉色边框

\newtcolorbox{prompt}[1]{
    enhanced,
    colback=bg-prompt,           % 使用淡粉背景，视觉更和谐
    colframe=s1_frame,          % 你的粉色边框
    fonttitle=\bfseries,       
    title={#1},
    fontupper=\small\ttfamily, % 关键：等宽字体
    arc=3pt,                   
    boxsep=4pt,
    left=4pt, right=4pt, top=4pt, bottom=4pt,
    attach boxed title to top left={xshift=2mm, yshift=-2mm},
    coltitle=black,
    boxed title style={colback=white, colframe=s1_frame, arc=2pt} 
}
\newcommand{\pvar}[1]{\textcolor{code_var}{\textbf{#1}}}
\newcommand{\phead}[1]{\vspace{0.4em}\noindent\textbf{#1}\vspace{0.2em}\newline}
\newcommand{\ptag}[1]{\textcolor{code_tag}{#1}}

\usepackage{microtype}
\usepackage{xspace}
\usepackage{booktabs}  % 用于三线表
\usepackage{tabularx}  % 用于自适应宽度
\usepackage{ragged2e}  % 用于单元格内自动换行对齐
\usepackage{amssymb}  % 用于 \checkmark 符号
\usepackage{multirow}
\usepackage{inconsolata}
\usepackage{pifont} % For checks and crosses
\usepackage{colortbl}
\usepackage{algorithm}
\usepackage{algpseudocode}
\definecolor{lightred}{RGB}{255, 230, 230}
\definecolor{lightgreen}{RGB}{230, 255, 230}
\definecolor{darkred}{RGB}{180, 0, 0}
\definecolor{darkgreen}{RGB}{0, 100, 0}
\definecolor{myGreen}{HTML}{008000} % 根据您的喜好定义具体绿色
\newcommand{\cmark}{\textcolor{myGreen}{\ding{51}}}
\newcommand{\xmark}{\textcolor{darkred}{\ding{55}}}%

\usepackage{graphicx}
\usepackage{url}

\newcommand{\eg}{\emph{e.g.,}\xspace}

\newcommand{\etc}{\textit{etc}\xspace}

\newcommand{\paratitle}[1]{\noindent\textbf{#1}}

\title{Beyond Static Dialogues: Benchmarking Realistic, Heterogeneous, and Evolving Long-Term Memory}

\author{
  \textbf{Han Zhang\textsuperscript{1,2,3}\thanks{Work done during internships at Microsoft.}},
  \textbf{Zihao Tang\textsuperscript{3}},
  \textbf{Xin Yu\textsuperscript{3}\thanks{Correspondence to Xin Yu and Hanfang Yang. \faEnvelope: \texttt{xinyu2@microsoft.com}, \texttt{hyang@ruc.edu.cn}.}},
  \textbf{Xiao Liu\textsuperscript{3}},
  \textbf{Yeyun Gong\textsuperscript{3}},
\\
  \textbf{Haizhen Huang\textsuperscript{3}},
  \textbf{Yan Lu\textsuperscript{3}},
  \textbf{Weiwei Deng\textsuperscript{3}},
  \textbf{Feng Sun\textsuperscript{3}},
  \textbf{Qi Zhang\textsuperscript{3}},
  \textbf{Hanfang Yang\textsuperscript{1,2}\footnotemark[2]}
\\
\\
  \textsuperscript{1}Center for Applied Statistics, Renmin University of China \\
  \textsuperscript{2}School of Statistics, Renmin University of China \\
  \textsuperscript{3}Microsoft
}

\begin{document}
\maketitle
\begin{abstract}
In existing memory benchmarks for Large Language Models (LLMs), the evaluated dialogue sessions often lack long-term semantic consistency, and the underlying personas tend to be flat and static. Furthermore, in real-world scenarios, interactions between users and assistants involve more diverse, heterogeneous data streams, such as documents and emails. These shortcomings significantly limit the realism and effectiveness of current evaluations. To address these limitations, we introduce \textbf{RHELM} (\underline{R}ealistic, \underline{H}eterogeneous, and \underline{E}volving \underline{L}ong-term \underline{M}emory). Driven by meticulously crafted user profiles and a novel \textbf{LOOP} (p\underline{L}an-r\underline{O}llout-ev\underline{O}lve-\underline{P}rune) module, we construct realistic dialogues across diverse interaction scenarios that exhibit dynamic temporal evolution and long-term coherence. Crucially, these dialogues are deeply integrated with heterogeneous external sources synchronized with the user's temporal event trajectory. The resulting benchmark encompasses challenging question-answer pairs spanning \textbf{seven} inquiry types, with each question mapping to at least one of \textbf{27} critical memory characteristics that we identify as essential yet underexplored in current research. Comprehensive experiments across full-context models, retrieval-augmented generation (RAG) methods, and representative memory frameworks reveal that contemporary approaches still expose critical weaknesses in complex, real-world settings, particularly in resolving multi-source aggregation and real-world contextual reasoning. The data is released at \url{https://github.com/microsoft/RHELM}.

\end{abstract}

\section{Introduction}

Recently, research on the memory capabilities of Large Language Models (LLMs) has garnered unprecedented attention~\cite{hu2025memory}. While scaling model size and extending context windows have enabled models to "memorize" vast amounts of general knowledge~\cite{hendrycks2020measuring, wang2024mmlu}, they often fail to satisfy the memory requirements of authentic personal assistant interactions. Perspectives from cognitive science~\cite{riedelDeclarativeMemory2015} point out that memory is intrinsically linked to personal traits, evolves dynamically over time, and depends on an individual's unique historical context. Regrettably, current general-purpose models and LLM agents remain limited in effectively capturing these nuanced attributes~\cite{wei2025evo}.

Although several benchmarks have emerged to evaluate the memory capabilities of conversational assistants, they continue to exhibit deficiencies in modeling real-world complexities in the following aspects~\cite{zhang2025survey,yehudai2025survey}:

\textbf{Absence of Semantic Coherence and Behavioral Fidelity.} Existing benchmarks frequently construct long-context history by inserting semantically disjoint conversational fillers. Consequently, both the dialogue segments and the underlying behavioral logic fail to form meaningful connections with the broader context~\cite{longmemo, jiang2025know}. Moreover, users engage in diverse communicative intents in real world-ranging from functional tasks to emotional disclosures. They communicate in varied ways, each reflecting a distinct user persona and varying levels of informational granularity. Consequently, evaluations based on simplistic synthetic dialogues lack realism, precluding a rigorous assessment of coherent memory.
\begin{figure*}[ht]
    \centering
    \vskip -0.2in
    \includegraphics[width=1\linewidth]{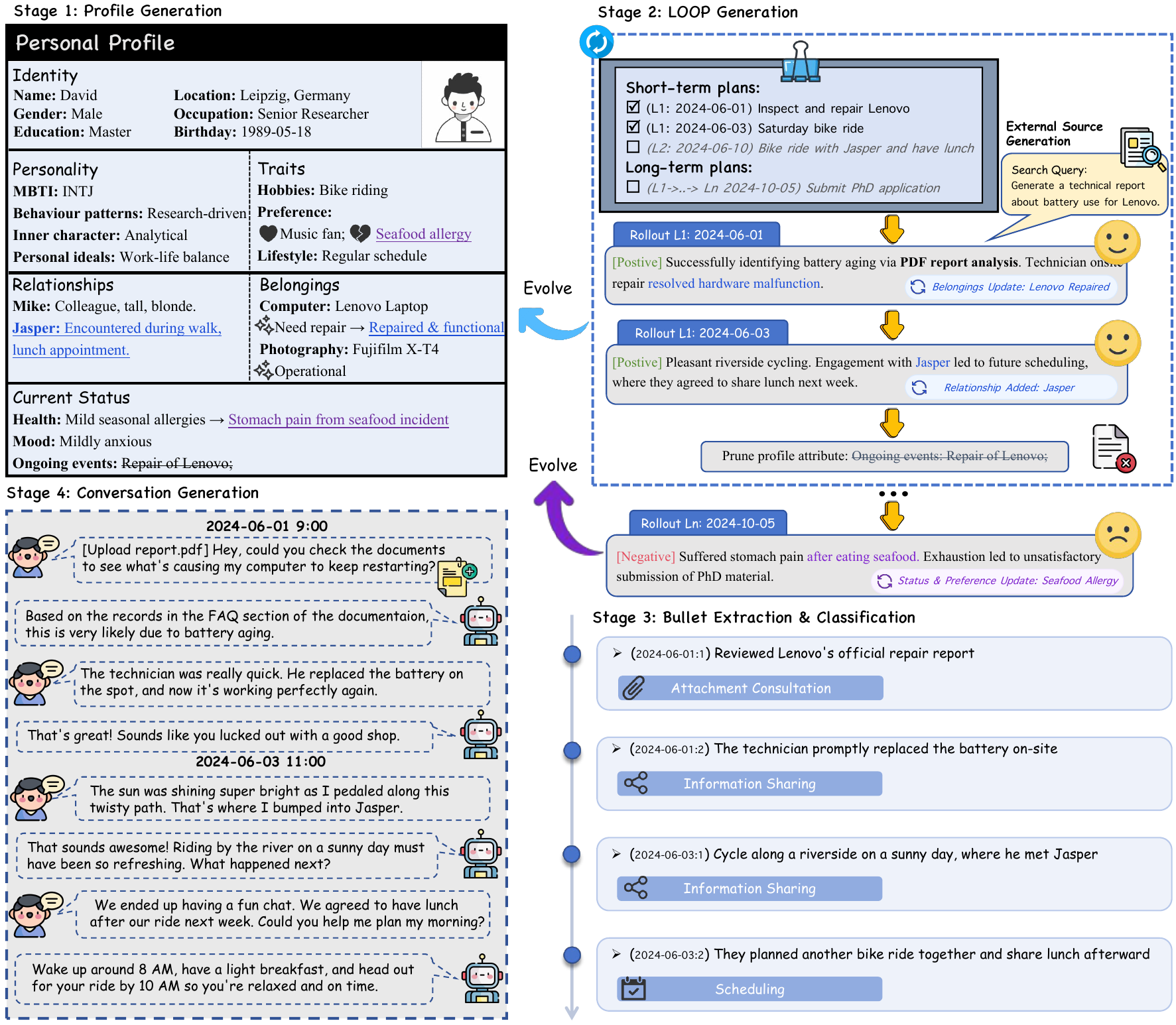}
    \caption{The overview of the \textbf{RHELM} benchmark curation.}
    \label{main}
    \vskip -0.2in
\end{figure*}
\begin{table*}[t]
\centering
% \small

\renewcommand{\arraystretch}{1.2}

\definecolor{oursbg}{gray}{0.9}
\vskip -0.1in
\caption{Comparison of representative AI memory benchmarks evaluated across multiple dimensions. Statistics are either from the original paper or based on our estimations.}
\label{tab:memory_benchmarks}
\vskip -0.1in

\resizebox{\textwidth}{!}{%
\begin{tabular}{l | c c c | c c c | c}
\toprule
\multirow{2}{*}{\textbf{Benchmark}} & \multicolumn{3}{c|}{\textbf{Benchmark Statistics}} & \multicolumn{3}{c|}{\textbf{Core Characteristics}} & \multirow{2}{*}{\begin{tabular}[c]{@{}c@{}}\textbf{Communication} \\ \textbf{Topic}\end{tabular}} \\
\cmidrule(lr){2-4} \cmidrule(lr){5-7}
 & \textbf{Avg. Turns} & \textbf{Context Window} & \textbf{External Data} & \textbf{Profile Evolution} & \textbf{Semantic Consistency} & \textbf{Misleading Test} & \\
\midrule

LoCoMo~\cite{maharana2024evaluating}    & $4.77$  & $10k$        & \xmark & \xmark & \cmark & \xmark & Daily Life \\
PerLTQA~\cite{du2024perltqa}            & $5.1$   & $1M$         & \xmark & \xmark & \xmark & \xmark & Daily Life \\
LongMemEval~\cite{longmemo}             & $5.19$  & $115k, 1.5M$ & \xmark & \xmark & \xmark & \xmark & Daily Life \\
PersonaMem~\cite{jiang2025know}         & $15-30$ & $128k-1M$    & \xmark & \cmark & \xmark & \xmark & Task-oriented \\
MT-Mind2Web~\cite{deng-etal-2024-multi} & $5.62$  & $\sim 400k$  & \cmark & \xmark & \xmark & \xmark & Web browsing \\
DialSim~\cite{kim2024dialsim}           & $12$    & $\sim 350k$  & \xmark & \xmark & \cmark & \xmark & TV shows \\
% MemBench~\cite{tan2025membench}         & $\sim 10$  & $10k$-$100k$ & \xmark & \cmark & \xmark & \xmark & Daily Life \\
% BEAM~\cite{tavakoli2025beyond}          & $107$-$7.8k$\textsuperscript{\dag} & $128k$-$10M$ & \xmark & \cmark & \cmark & \cmark & Multi-domain \\

\midrule

\rowcolor{oursbg}
\textbf{RHELM (Ours)}                   & \textbf{18.7} & \textbf{500k-1M} & \textbf{\cmark} & \textbf{\cmark} & \textbf{\cmark} & \textbf{\cmark} & \textbf{Multi-faceted} \\
\bottomrule
\end{tabular}%
}

\renewcommand{\arraystretch}{1}
\end{table*}
\begin{table}[htbp]
    \centering
    \small
    \caption{Overview statistics of the \textbf{RHELM} Dataset.}
    \label{tab:dataset_summary_compact}
    \begin{tabularx}{\columnwidth}{@{}Xr@{}}
        \toprule
        \textbf{Metric} & \textbf{Statistics} \\
        \midrule
        \multicolumn{2}{@{}l}{\textit{\textbf{Dialogue History}}} \\
        \quad Total Days (Sessions) & 629 \\
        \quad Total Turns & 11,764 \\
        \quad Total Dialogue Tokens & 4,776,002 \\
        \midrule
        \multicolumn{2}{@{}l}{\textit{\textbf{External Sources}}} \\
        \quad Total External Files & 2,180 \\
        \quad Total File Tokens & 2,428,121 \\
        \midrule
        \multicolumn{2}{@{}l}{\textit{\textbf{Evaluation Data}}} \\
        \quad Total QA Pairs & 1,305 \\
        \bottomrule
    \end{tabularx}
    \vskip -0.1in
    \vskip -0.1in
\end{table}

\textbf{Homogeneous Information Sources.} Existing AI assistants are evolving from basic chatbots to systems capable of reasoning across diverse data sources~\cite{comanici2025gemini}. Sources such as reports, journals and emails serve as rich repositories of both episodic and semantic information~\cite{lei-etal-2023-unsupervised}. Nevertheless, most current benchmarks are largely confined to conversational interactions. In practice, a robust AI assistant must synthesize information across heterogeneous sources, ranging from unstructured text and semi-structured tables to web pages~\cite{li2024personal}. While distinct from colloquial text, these structured artifacts possess high information density and user-specific details essential for comprehensive memory formation, posing a significant challenge to the modeling of user-centric memory systems.

\textbf{Omission of Memory-Conditioned Misleading Queries.} In existing benchmarks, the majority of queries demand explicit factual answers, treating LLMs as static retrieval systems under the "Needle-in-a-Haystack" paradigm~\cite{needle-in-haystack}. While some frameworks~\cite{tavakoli2025beyond} attempt to evaluate robustness by directly injecting hallucinations into questions, they still overlook a critical challenge in real-world scenarios: implicit state constraints. Real-world users may propose authentic requests that contradict their own grounded realities, past events, or evolving preferences. A truly capable memory-augmented assistant should not act merely as a "pure instruction-follower". Instead, it must proactively track the user's implicit status from historical interactions, detect conflicting requests, and respond proactively based on the user's authentic condition.

To bridge these gaps, we introduce \textbf{RHELM} (\underline{R}ealistic, \underline{H}eterogeneous, and \underline{E}volving \underline{L}ong-term \underline{M}emory), a benchmark designed to comprehensively evaluate the complex memory capabilities of personal AI assistants. Unlike prior statically assembled datasets, \textbf{RHELM} models consistent and realistic long-term user behaviors through a dynamic \textbf{LOOP} (\textsl{p\underline{L}an, r\underline{O}llout, ev\underline{O}lve, and \underline{P}rune}) module, grounded in meticulously designed personas. Furthermore, rather than relying exclusively on daily dialogues, we synthesize heterogeneous external sources (\eg reports, journals, emails) via Deep Research methodologies~\cite{openai2025deepresearch}. Operating on timelines spanning a simulated one-year period, \textbf{RHELM} ultimately yields \textbf{10} distinct persona trajectories encompassing \textbf{11,764} turns and \textbf{2,180} external sources with context lengths ranging from 500k to 1M tokens in total per persona. For the evaluation suite, we construct \textbf{1,305} cognitively demanding question-answer pairs spanning \textbf{seven} categories and \textbf{27} distinct complexity features, including a novel category specifically targeting the aforementioned misleading dimension to facilitate implicit status reasoning. Table~\ref{tab:memory_benchmarks} compares our dataset against existing benchmarks, while Table~\ref{tab:dataset_summary_compact} summarizes \textbf{RHELM}'s statistics (details in Appendix~\ref{app: statistics}).

In summary, our main contributions are as follows: (\romannumeral1) We present \textbf{RHELM}, a highly realistic benchmark for long-horizon memory evaluation, uniquely bridging the gap between conversational interactions and heterogeneous external data streams. (\romannumeral2) We define a systematic evaluation taxonomy encompassing \textbf{27} challenging memory characteristics that we identify as essential yet underexplored in current research. (\romannumeral3) Comprehensive experiments across three memory-augmented settings demonstrate that even state-of-the-art methods still struggle in real-world memory reasoning, highlighting a clear path for future enhancements.

\section{Related Work}
\paratitle{Long-context LLMs \& Memory Mechanisms.}
Recent strides in LLMs have significantly expanded their context windows as working memory. This progress is underpinned by efficient attention mechanisms~\cite{Beltagy2020Longformer,kwon2023efficient} that reduce computational overhead, and advanced techniques applied during fine-tuning~\cite{su2024roformer,peng2023yarn,press2021train} to facilitate length extrapolation. Consequently, modern proprietary models now support massive context windows tailored for complex tasks~\cite{anthropic2025claude4_systemcard,comanici2025gemini}. Beyond direct context extension, more sophisticated memory systems enhance the management of authentic memory scenarios by effectively compressing and organizing historical information~\cite{packer2023memgpt,chhikara2025mem0}.

\paratitle{Long-term Dialogue Benchmarks.}
As the demand for sophisticated memory capabilities has progressively intensified, benchmarks have transitioned from text-centric evaluations such as LongBench~\cite{bai2024longbench, longbenchv2} toward conversational frameworks like LoCoMo~\cite{maharana2024evaluating}. DialSim~\cite{kim2024dialsim} prioritizes role-playing to render dialogues more authentic. PerLTQA~\cite{du2024perltqa} emphasizes the capture of social networks and semantic information, while PersonaMem~\cite{jiang2025know} necessitates tracking shifts in user preferences. Furthermore, LongMemEval~\cite{longmemo} and BEAM~\cite{tavakoli2025beyond} extend this further by querying the most recent evolved details associated with users with longer context. Despite these advancements, the integration of heterogeneous data sources and the modeling of realistic, dynamic user trajectories remain underexplored.

\section{Overview}
\subsection{Problem Formulation}

We formally define the task of our memory evaluation as follows. Consider a persona $P$, associated with a time span $[\tau_s, \tau_e]$. Within this period, the user's historical context is composed of two heterogeneous streams: conversational dialogues and external data sources. We denote the \textbf{Dialogue Stream} as $\mathcal{C} = \{(\tau_i, x_i, y_i)\}_{i=1}^{N}$, where each tuple represents a dialogue turn occurring at timestamp $\tau_i$, consisting of a user utterance $x_i$ and an assistant response $y_i$. Parallel to the dialogue, we define the \textbf{External Source Stream} as $\mathcal{E} = \{(\tau_j, d_j)\}_{j=1}^{M}$, where $d_j$ represents a textual data chunk (\eg a document fragment) available at timestamp $\tau_j$. During the evaluation phase, a query $q$ is issued at a specific query time $\tau_q$. The objective of the model is to generate an answer $a$ based on all the information available. The expected answer $a$ can manifest either as a concise phrase or a descriptive natural language response—for instance, identifying and rectifying a conflict within the user's query. 

\subsection{RHELM Overview}

We define a taxonomy of seven core query categories, as illustrated in Table~\ref{tab:memory_taxonomy}, encompassing five dialogue-centric types—namely \emph{Fact}, \emph{Temporal}, \emph{Hallucination}, \emph{Aggregation} and \emph{Misleading}, and two types with heterogeneous sources: \emph{External Source} queries and \emph{Mixed} queries. While certain categories, such as \emph{Fact} and \emph{Temporal}, have been partially explored in prior benchmarks~\cite{longmemo,maharana2024evaluating}, we significantly extend the complexity across all seven categories to facilitate a more rigorous evaluation of model memory. Specifically, we propose \textbf{27} core challenging features in total. These features necessitate deeper levels of reasoning, sustained tracking of long-horizon dependencies, and multifaceted aggregation of information across heterogeneous sources. During the construction of question-answer pairs, the feature definitions are integrated to ensure that each query encapsulates at least one such characteristic. To facilitate efficient evaluation, we ensure that most questions have short phrases as answers. In particular, \emph{Hallucination} and \emph{Misleading} types require models not only to identify false claims but also to specify the correct factual context.
\begin{table}[t!]
\centering
\scriptsize
\vskip -0.1in
\caption{\small Taxonomy of challenging questions. Both Attachment and Email correspond to the \textit{External Source} type. More detailed definitions are listed in the Table~\ref{tab:memory_taxonomy_app}.}
\vskip -0.05in
\label{tab:memory_taxonomy}
\setlength{\tabcolsep}{2pt}
\renewcommand{\arraystretch}{1.10}
\newcommand{\twocell}[2]{%
\begin{tabular}{@{}>{\raggedright\arraybackslash}p{0.47\linewidth}@{\hspace{0.04\linewidth}}>{\raggedright\arraybackslash}p{0.47\linewidth}@{}}%
#1 & #2%
\end{tabular}}
\newcommand{\onecell}[1]{%
\begin{tabular}{@{}>{\raggedright\arraybackslash}p{\linewidth}@{}}%
#1%
\end{tabular}}
\begin{tabularx}{\columnwidth}{@{}>{\bfseries\raggedright\arraybackslash}p{0.24\columnwidth} >{\raggedright\arraybackslash}X@{}}
\toprule
\textbf{Category} & \textbf{Characteristics} \\
\midrule
\rowcolor{lightgray}
\multicolumn{2}{c}{\textbf{I. Dialogue History QA}} \\
\addlinespace[1pt]
Fact & \twocell{Multi-Hop Traversal}{Entity Disambiguation} \\
     & \twocell{State-Dependent Attribute}{Negative Constraints} \\
\addlinespace[1pt]\cmidrule(lr){1-2}\addlinespace[1pt]
Temporal & \twocell{Indirect Identification}{Sequence Comprehension} \\
	   & \twocell{Long-Horizon Synthesis}{Implicit Temporal Lookup} \\
\addlinespace[1pt]\cmidrule(lr){1-2}\addlinespace[1pt]
Hallucination & \twocell{Misattribution}{Fabrication} \\
		  & \twocell{Preference Conflict}{Contextual Contradiction} \\
\addlinespace[1pt]\cmidrule(lr){1-2}\addlinespace[1pt]
Aggregation & \twocell{Conditional Counting}{Trend Analysis} \\
		& \twocell{Extreme Value}{Absence Detection} \\
\addlinespace[1pt]\cmidrule(lr){1-2}\addlinespace[1pt]
Misleading & \twocell{Implicit State Conflict}{Proactive Response} \\
\midrule
\rowcolor{lightgray}
\multicolumn{2}{c}{\textbf{II. External Source QA}} \\
\addlinespace[1pt]
Attachment & \twocell{Fact Retrieval}{Table Reasoning} \\
	     & \twocell{Structural Navigation}{Table Aggregation} \\
\addlinespace[1pt]\cmidrule(lr){1-2}\addlinespace[1pt]
Email & \twocell{Cross-time Count}{Email Localization} \\
\midrule
\rowcolor{lightgray}
\multicolumn{2}{c}{\textbf{III. Hybrid Context QA}} \\
\addlinespace[1pt]
Mixed & \twocell{Relative Positioning}{Contextual Retrieval} \\
	& \onecell{Post-Modification Analysis} \\
\bottomrule
\end{tabularx}
\vskip -0.15in
\end{table}

\section{Benchmark Curation}
This section details the systematic process of constructing the benchmark. The overall curation pipeline is illustrated in Figure~\ref{main}, and the main workflow is outlined in Algorithm~\ref{alg:curation}.

\subsection{Profile Generation}
A rich and deeply layered character is pivotal to the entire creation process for \textbf{RHELM}. As users often exhibit dynamic evolution in factual details while maintaining consistency within their core qualities throughout lifelong learning, we developed a six-dimensional persona taxonomy. These attributes range from internal psychology to external realities, and from immutable characteristics to transient states: \emph{Identity}, \emph{Personality}, \emph{Traits}, \emph{Relationships}, \emph{Belongings}, and \emph{Current Status}. Throughout the benchmark generation process, these profiles are dynamically updated and refined. To ensure the accuracy and integrity of these updates, profiles are stored following a rigorous JSON schema, in which each attribute is governed by strict definitions and pre-defined data types. A sample profile and an overview of the personas used in \textbf{RHELM} are provided in Appendix~\ref{app: profile}.

\subsection{LOOP (p\underline{L}an-r\underline{O}llout-ev\underline{O}lve-\underline{P}rune)}
\begin{algorithm}[t]
\caption{\textbf{RHELM} Workflow.}
\label{alg:curation}
\small
\newcommand{\loopstep}[1]{\textit{\textcolor{gray!85!black}{#1}}}
\begin{algorithmic}[1]
\Require initial persona $P$; time span $[\tau_{s}, \tau_{e}]$; rollout probability $p$; prune schedule $\rho$
\Ensure profile trajectory $\mathcal{P}$, external sources $\mathcal{E}$, dialogues $\mathcal{C}$
\State \textcolor{gray}{\textbf{\# Profile Generation}}
\State $\mathbf{P}_{\tau_{s}} \gets \Call{EnrichProfile}{P}$ \Comment{6-dim JSON profile}
\State \textcolor{gray}{\textbf{\# LOOP Module}}
\State $\tau \gets \tau_{s}$
\While{$\tau \leq \tau_{e}$}
    \State $(\mathbf{g}_{\tau'}, \tau') \gets \Call{Plan}{\mathbf{P}_{\tau}, \tau}$ \Comment{\loopstep{pLan}}
    \State $\mathbf{o}_{\tau'} \gets \Call{Rollout}{\mathbf{g}_{\tau'}, \tau', p}$ \Comment{\loopstep{rOllout}}
    \State $\mathcal{E}_{\tau'} \gets \Call{ExternalGen}{\mathbf{o}_{\tau'}}$
    \State $\mathbf{P}_{\tau'} \gets \Call{Evolve}{\mathbf{P}_{\tau}, \mathbf{o}_{\tau'}}$ \Comment{\loopstep{evOlve}}
    \If{$\Call{ShouldPrune}{\tau', \rho}$}
        \State $\mathbf{P}_{\tau'} \gets \Call{Prune}{\mathbf{P}_{\tau'}}$ \Comment{\loopstep{Prune}}
    \EndIf
    \State $\tau \gets \tau'$
\EndWhile
\ForAll{simulated date $\tau$}
    \State \textcolor{gray}{\textbf{\# Bullet Extraction \& Classification}}
    \State $\mathbf{b}_{\tau} \gets \Call{BulletExtract}{\mathbf{o}_{\tau}}$
    \State $\mathbf{r}_{\tau} \gets \Call{Classify}{\mathbf{b}_{\tau}}$ \Comment{Five dialogue categories}
    \State \textcolor{gray}{\textbf{\# Dialogue Generation}}
    \State $\mathcal{C}_{\tau} \gets \Call{DialogueGen}{\mathbf{b}_{\tau}, \mathbf{r}_{\tau}, \mathbf{P}_{\tau}, \mathcal{E}_{\tau}}$
\EndFor
\State \Return $\{\mathcal{P}, \mathcal{E}, \mathcal{C}\}$
\end{algorithmic}
\end{algorithm}

The LOOP module simulates realistic lifelong trajectory. Leveraging specific user profile, the model generates \textbf{\emph{plans}} encompassing both short-term arrangements (social interactions, routines, and personal interests) and long-term projections (career progression, life milestones, and significant personal transitions). For each scheduled event, we utilize a \textbf{\emph{rollout}} mechanism controlled by a probability $p$, yielding either positive or negative outcomes. The outcomes comprise detailed event narratives about the day. Empirically, we observe that this simple mechanism effectively simulates the fluctuations and contingencies in real life. Based on these outcomes, the model dynamically \textbf{\emph{evolves}} the previous profile to reflect current state changes. This update process is implemented through functional calls on JSON schema. Furthermore, to mitigate the risk of cumulative semantic drift or error propagation over extended temporal horizons, a \textbf{\emph{prune}} module is employed. This module periodically recalibrates the user profile and prunes outdated entities. Following each pruning iteration, a new LOOP cycle is re-initialized, ensuring consistent long-term update through the user's life trajectory.

\paratitle{Outcome Rollout.} To simulate the stochasticity of external contingencies in real-world environments, we introduce a random factor $p$ to govern the valence—\textit{positive} or \textit{negative}—of event trajectories. Negative outcomes, which denote life failures or unforeseen accidents, serve as critical milestones that significantly alter the evolution of the narrative arc. For instance, a physical injury incurred during a cycling excursion may fundamentally influence subsequent scheduled activities. By incorporating such perturbations, we facilitate the development of long-tail event sequences~\cite{li-etal-2024-search}, enabling the evaluation of model performance in complex, non-linear scenarios that are difficult for current models to capture.

\paratitle{External Sources.} Contemporary application scenarios for personal AI assistants no longer rely solely on conversational interaction. Consequently, incorporating complex heterogeneous sources is essential for the construction of robust benchmarks. To mirror the multi-faceted information users encounter in daily life, we synthesize diverse external data sources conditioned on the daily outcome narratives. Specifically, we focus on three primary categories: emails, personal journals, and professional reports, which manifest in various formats including text, markdown, HTML. To ensure these artifacts possess high degrees of reality and complexity, we leverage Deep Research methodologies~\cite{langchain2025opendeep, openai2025deepresearch} to generate the latter two data sources. Based on the outcomes generated during the LOOP phase, the model is tasked with creating research queries for reports and journals relevant to the user's day; these queries are subsequently processed by a deep research agent. In the final stage of generation, the presentation of the outputs is further refined to reflect the formal professionalism of reports or the intimate, personal nature of journals. Detailed statistical reports are available in the Appendix~\ref{app: statistics}.

\paratitle{Profile Evolution.}
To maintain dynamically evolving user states, we continuously update the initialized profiles based on daily outcomes via a dual-phase paradigm: \textit{factual evolution} and \textit{state evolution}. The former extracts fine-grained details from events to revise objective attributes (\eg, social relationships and belongings), while the latter infers intrinsic shifts, such as evolving preferences and hobbies. This disentangled strategy enables the model to independently capture both external and internal user dynamics. 

The dynamic updates primarily involve adding of new items, modifying existing ones, or removing outdated items. By enforcing rigorous type definitions and schema constraints on profile attributes, we ensure the structural integrity and validity of the profile evolution throughout the execution. We further provide a detailed analysis for the evolution frequency in Appendix~\ref{app:profile_evolution}.

\subsection{Dialogue Synthesis}
Below we introduce the dialogue construction process using the trajectory from the LOOP module. 

\paratitle{Bullet Extraction \& Classification.}
Given the outcome narratives $\mathbf{o}_{\tau}$, we decompose them into atomic bullet points $\mathbf{b}_{\tau}$, ensuring each capture essential details. These bullets are then classified into different dialogue categories based on the user's likely communicative intent and emotional context. Specifically, we introduce a taxonomy of five dialogue categories: \textit{information sharing}, \textit{advice seeking}, \textit{status updates}, \textit{scheduling}, and \textit{attachment consultation}. This categorization mirrors real-world dynamics where a user's tone and communication style adapt to their needs; for instance, document-centric discussions typically exhibit greater professionalism and verbosity than routine factual exchanges. 

\paratitle{Dialogue Generation.} 
Each bullet serves as a thematic anchor for a specific topic. To naturally form dialogue streams, we design two dialogue modes per topic: an \textit{initial turn}, where the user naturally leads in a new conversational thread, and a \textit{follow-up turn}, where the user continues an ongoing discussion. Different categories further adopt specific interaction patterns and individual user personas.  Exemplary conversation and more implementation details are provided in Appendix~\ref{app:conversation}. 

\begin{table*}[t]
    \vspace{-5mm}
    \caption{\small Detailed Performance Evaluation on \textbf{RHELM}. The two evaluation settings (With / Without External Data Sources) are presented side-by-side. The evaluation metrics are grouped conceptually into Dialogue History QA (\textbf{FC}: Fact, \textbf{TP}: Temporal, \textbf{AG}: Aggregation, \textbf{HL}: Hallucination,   \textbf{MI}: Misleading), External Source QA (\textbf{EX}: Attachment and Email), and Hybrid Context QA (\textbf{MX}: Mixed). Overall best scores are marked in \textbf{bold}, and second-best scores are \underline{underlined}.}
    \centering
    % \vskip -1.1em
    \resizebox{1.0\linewidth}{!}{
    \renewcommand{\arraystretch}{1.2}
    \setlength{\tabcolsep}{4pt}
    \begin{tabular}{l|ccccc|c|c|c|ccccc|c|c|c}
        \toprule
        \multirow{2.5}{*}{\bf Model} &
        \multicolumn{8}{c|}{{\cellcolor{green!10}}{\bf Without External Data Sources}} &
        \multicolumn{8}{c}{{\cellcolor{blue!10}}{\bf With External Data Sources}}\\

        \cmidrule(lr){2-9} \cmidrule(lr){10-17}

        & \textbf{FC} & \textbf{TP} & \textbf{AG} & \textbf{HL} & \textbf{MI} & \textbf{EX} & \textbf{MX} & \textbf{Avg}
        & \textbf{FC} & \textbf{TP} & \textbf{AG} & \textbf{HL} & \textbf{MI} & \textbf{EX} & \textbf{MX} & \textbf{Avg} \\
        \midrule

        % ==========================================
        % RAG Baselines
        % ==========================================
        \rowcolor{gray!15}\multicolumn{17}{c}{\emph{RAG Baselines}}  \\
        \bf GPT-4.1-mini ($k=5$) &
        35.8 & 17.3 & 17.7 & 15.2 & 3.1 & 8.0 & 10.0 & 16.3 &
        34.8 & 14.1 & 19.3 & 11.7 & 3.1 & 16.9 & 12.4 & 17.5 \\

        \bf GPT-4.1-mini ($k=20$) &
        44.0 & 32.4 & 31.8 & 18.3 & 3.1 & 12.1 & 12.9 & 23.5 &
        42.5 & 28.7 & 30.7 & 13.2 & 3.1 & 28.5 & 13.8 & 25.1 \\

        \bf GPT-4.1-mini ($k=50$) &
        \textbf{59.9} & \textbf{41.6} & \textbf{40.1} & 15.7 & 1.5 & 12.9 & 16.7 & 28.9 &
        \textbf{54.6} & \textbf{39.5} & \textbf{38.0} & 11.2 & 1.5 & \underline{38.6} & 22.4 & 32.6 \\

        \bf Hybrid ($k=5$) &
        34.3 & 20.5 & 14.1 & 19.8 & 1.5 & 8.0 & 10.5 & 16.7 &
        31.9 & 19.5 & 14.1 & 19.3 & 1.5 & 16.1 & 10.5 & 17.6 \\

        \bf Hybrid ($k=20$) &
        47.3 & 35.7 & 31.8 & 19.3 & 3.1 & 10.4 & 15.2 & 24.8 &
        45.9 & 30.8 & 26.6 & 16.8 & 4.6 & 26.9 & 15.7 & 26.0 \\

        \bf Hybrid ($k=50$) &
        \underline{56.5} & \underline{41.1} & \underline{35.9} & 15.2 & 3.1 & \underline{13.7} & 16.7 & 27.8 &
        \underline{53.1} & \underline{37.8} & 33.9 & 8.6 & 3.1 & 33.3 & 18.6 & 29.6 \\

        \bf GPT-4.1 ($k=20$)&
        51.7 & 34.1 & \underline{35.9} & 23.9 & 7.7 & \textbf{16.1} & \underline{17.6} & 28.2 &
        50.2 & 29.2 & 32.3 & 19.8 & 6.2 & 32.5 & 19.5 & 29.5 \\

        \bf Gemini-2.5-Pro ($k=20$) &
        45.4 & 35.1 & 27.1 & \underline{66.0} & \underline{23.1} & 12.4 & \textbf{18.1} & \underline{32.6} &
        43.0 & 31.9 & 26.0 & \textbf{64.5} & \underline{26.2} & 31.3 & 20.5 & \underline{35.5} \\

        \bf Claude-Opus-4.5 ($k=20$) &
        50.7 & 37.8 & 33.3 & \textbf{68.0} & \textbf{47.7} & \underline{13.7} & 16.2 & \textbf{36.2} &
        50.2 & 30.8 & 31.8 & \underline{60.9} & \textbf{41.5} & 33.7 & 21.0 & \textbf{38.1} \\

        % ==========================================
        % Long-Context Models
        % ==========================================
        \rowcolor{blue!5}\multicolumn{17}{c}{\emph{Long-Context Models}} \\
        \bf Gemini-2.5-Flash-Lite (1M) &
        33.2 & 22.7 & 15.2 & 17.3 & 0.0 & 9.5 & 5.6 & 16.0 &
        31.7 & 14.1 & 23.4 & 7.6 & 0.0 & 19.0 & 13.1 & 17.3 \\
        \bf Qwen-2.5-14B-Instruct (1M) &
29.5 & 15.1 & 29.7 & 3.1 & 0.0 & 11.7 & 9.1 & 15.3 &
16.9 & 7.0 & 15.6 & 1.0 & 0.0 & 5.2 & 6.2 & 8.1 \\
        \bf GPT-4.1-mini (1M) &
        55.1 & 31.9 & \textbf{40.1} & 4.1 & 1.5 & 11.2 & 12.4 & 24.0 &
        49.3 & 27.0 & 33.9 & 2.0 & 1.5 & \textbf{43.4} & 0.3 & 33.9 \\

        % ==========================================
        % Memory Frameworks
        % ==========================================
        \rowcolor{green!5}\multicolumn{17}{c}{\emph{Memory Frameworks}} \\
        \textbf{MemGPT}~\cite{packer2023memgpt} &
        31.9 & 18.4 & 22.9 & 0.5 & 0.0 & 7.6 & 8.1 & 13.9 &
        27.5 & 14.6 & 28.7 & 1.5 & 1.5 & 18.9 & 17.1 & 17.3 \\

        \textbf{Mem0}~\cite{chhikara2025mem0} &
        41.6 & 31.4 & 28.1 & 10.7 & 3.1 & 10.8 & 13.3 & 21.1 &
        46.4 & 29.2 & 27.1 & 10.2 & 3.1 & 31.3 & \underline{35.7} & 28.9 \\

        \textbf{MemU}~\cite{nevamind2025memu} &
        49.3 & 32.4 & 33.9 & 8.6 & 4.6 & 12.0 & 11.4 & 23.1 &
        \textbf{54.6} & 36.2 & \underline{35.4} & 10.2 & 3.1 & 36.5 & \textbf{36.7} & 33.6 \\

        \bottomrule
    \end{tabular}
    }
    \vspace{-1mm}

    \label{tab:main_results_final}
    \vspace{-4mm}
\end{table*}
\section{Question Curation}
During the question generation process, we employ diverse sampling strategies to extract consecutive, cross-day event bullets. We then synthesize complex questions by incorporating these evidence bullets, the formal definition of the target question category, and its associated challenging characteristics into the model (as detailed in Table~\ref{tab:memory_taxonomy_app}).

\paratitle{Memory-Conditioned Misleading Queries.}
We innovatively introduce \textit{Memory-Conditioned Misleading Queries}. Under this evaluation dimension, the assistant is required to be aware of the user's ongoing state or preferences from implicit user queries. To rigorously assess the capability, we utilize ``trap'' queries wherein the user proposes a request that directly conflicts with the implicit constraints imposed by their updated life state. During the generation of these queries, critical, life-altering events (\eg chronic injuries, residential relocations, or sudden career shifts)  are deliberately extracted as grounding evidence. 

When formulating the response, the assistant must not blindly follow the user's explicit instruction. Instead, it is expected to proactively retrieve the historical event, deduce the ongoing restriction, politely identify the implicit conflict, and synthesize a constraint-compliant alternative.

\paratitle{Verifier-Assisted Auditing.} To maintain rigorous quality control across the whole generative pipeline, we implement a comprehensive \emph{Verifier} system designed to audit the outputs of each distinct phase. In the context of long-range trajectory synthesis, human auditing is susceptible to "attention drift" \cite{zouhar-etal-2025-ai} and entails prohibitive labor costs. Consequently, we deploy a suite of stringent verification modules that span the entire curation lifecycle—from profile evolution, external source synthesis to dialogue and QA pair generation—to ensure semantic consistency and factual integrity. Our findings indicate that the auxiliary information generated by the verifier module serves as a potent diagnostic tool, aiding human review while significantly reducing the requisite manual overhead. More details can be found in Appendix~\ref{app:verifier}. 

\section{Experiments}
Through our experiments, we aim to answer the following research questions (RQs):

\noindent $\bullet$ \textbf{RQ1:} To what extent do heterogeneous external sources impact the performance of current memory paradigms, and how do these sources interact with dialogue history? \\
\noindent $\bullet$ \textbf{RQ2:} How robust are current models against more realistic challenges, particularly hallucination and misleading issues? \\
\noindent $\bullet$ \textbf{RQ3:} How effectively do existing retrieval methods recall relevant evidence from long-horizon, heterogeneous user histories? \\
\noindent $\bullet$ \textbf{RQ4:} Which specific challenging characteristics expose the most significant deficiency?

\subsection{Experimental Setup}

We evaluate three distinct memory paradigms under two experimental configurations: one incorporating external sources and another excluding them.

\paratitle{RAG Baselines.} We adopt distinct chunking strategies for the two data streams: dialogue histories are segmented by individual turns, while external documents are split into fixed-length chunks of 500 tokens with chunk overlap of 50 tokens. All chunks are encoded using bge-large-en-v1.5~\cite{bge_embedding} and indexed via FAISS~\cite{johnson2019billion} for efficient similarity search. At inference time, we retrieve the top-$k$ most relevant chunks with $k \in \{5, 20, 50\}$. Using GPT-4.1-mini as the default LLM, we further evaluate GPT-4.1, Gemini-2.5-Pro, and Claude Opus 4.5 at $k=20$. In addition, we implement a hybrid retrieval variant that combines dense retrieval with BM25~\cite{robertson2009probabilistic} sparse retrieval via reciprocal rank fusion (RRF)~\cite{cormack2009reciprocal}.

\paratitle{Long-Context Models.} We evaluate GPT-4.1-mini, Gemini-2.5-Flash-Lite, and Qwen2.5-14B-Instruct-1M~\cite{qwen2.5-1m} for full-context inference. All models support context windows of up to 1M tokens. The dialogue histories and external sources are concatenated in chronological order and provided as a single input context.

\paratitle{Memory Frameworks.} We further evaluate three representative memory-augmented frameworks. {MemGPT}~\cite{packer2023memgpt} implements an OS-inspired virtual memory hierarchy. {Mem0}~\cite{chhikara2025mem0} provides a graph-based and vector-based hybrid memory layer that automatically extracts, consolidates, and retrieves user-specific memories. \textbf{MemU}~\cite{nevamind2025memu} maintains a hierarchical memory architecture in which dialogues and documents are processed through separate pipelines. All three frameworks use GPT-4.1-mini as the backbone LLM to ensure a fair comparison.

\textbf{Evaluation Metrics.} Utilizing LLMs for evaluation has increasingly become an efficient approach for performance assessment~\cite{gu2024survey}. Specifically, for hallucination and misleading type, responses involve identifying correct contexts and proactively refusing; in such cases, traditional metrics like Exact Match or BLEU~\cite{papineni2002bleu} often exhibit significant deficiency. Consequently, we employ the LLM-as-judge paradigm~\cite{liu-etal-2023-g} throughout experiments. Human validation and the detailed prompt are listed in the Appendix~\ref{human-verify} and Figure~\ref{fig:prompt-llm-judge}.

\subsection{Experimental Results}

\begin{figure*}[htbp]
    \centering
    \vskip -0.4in
    \includegraphics[width=\linewidth]{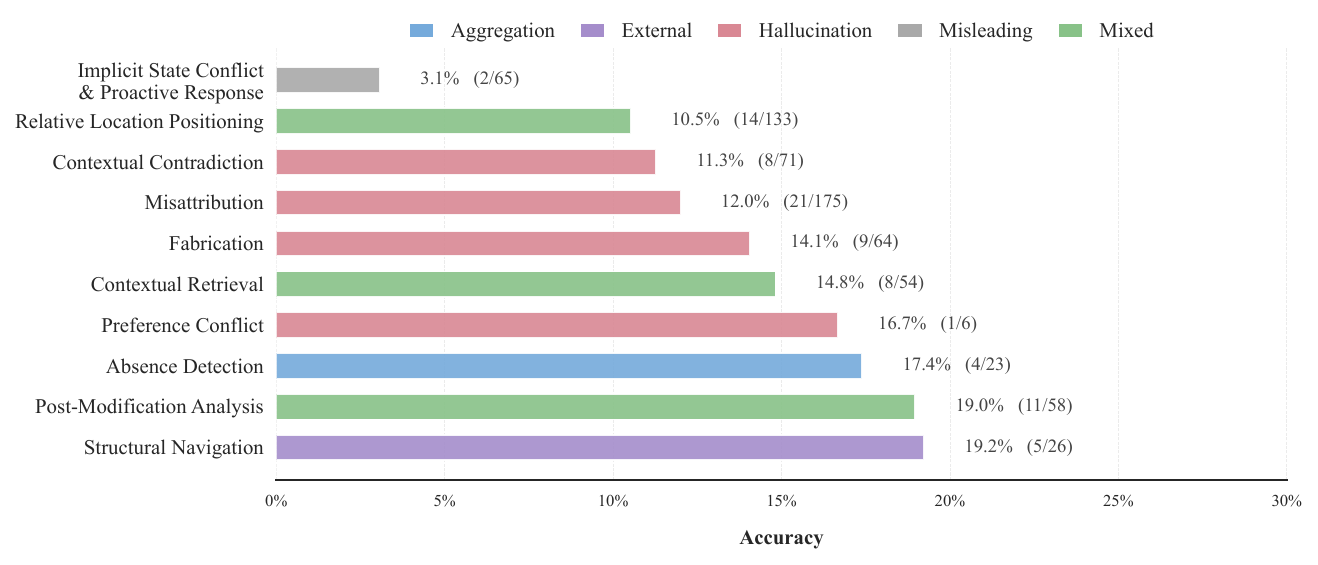}
    \vskip -0.1in
    \caption{Analysis of the 10 worst-performing challenging characteristics in \textbf{RHELM}. Models exhibit notably poor performance on features involving cross-source aggregation and real-world contextual reasoning.}
    \label{fig:worst_characteristics}
\end{figure*}
\paratitle{Main Results.} 
Table~\ref{tab:main_results_final} presents the main results. Further analysis yields several pivotal insights:

\textbf{RQ1:} The experimental results reveal that all three paradigms exhibit markedly constrained performance on the \textbf{RHELM} benchmark. The best-performing model, Claude Opus 4.5, achieves an average performance of only $38.1$ with external sources, and $36.2$ without them. Further analysis shows that models without external sources can partially address queries in \textbf{EX} and \textbf{MX} types, suggesting conversational histories can serve as auxiliary references regarding user life trajectory. However, introducing external sources introduces challenges for standard types (\eg performance decreases from $44.0$ to $42.5$ for RAG ($k=20$), and from $59.9$ to $54.6$ for RAG ($k=50$)). This highlights the necessity of integrating diverse data formats into the evaluation suite and memory systems. Notably, all models struggle considerably with mixed-type queries, particularly RAG-based approaches. This deficiency illuminates the inadequacy of isolated retrieval mechanisms in forging a unified memory architecture capable of cross-modal reasoning~\cite{wang-etal-2024-leave}.

\textbf{RQ2:} Models exhibit notably poor performance on hallucination and misleading queries, which more accurately reflect real-world scenarios. In RAG-based methods, increasing the volume of retrieved evidence further degrades performance (e.g., from $13.2$ to $11.2$ for hallucination type). Notably, nearly all methods fail severely on the misleading type, with accuracy falling below $5\%$. However, stronger reasoning models such as Claude Opus 4.5 and Gemini-2.5-Pro achieve substantially better results on these two types compared to other models, suggesting that enhanced reasoning capabilities enable models to more effectively detect and resist deceptive or fabricated premises. Nevertheless, how to effectively distinguish between seemingly plausible user trajectories and genuine user states remains an open challenge.

\begin{figure}[htbp]
    \centering
    \includegraphics[width=\linewidth]{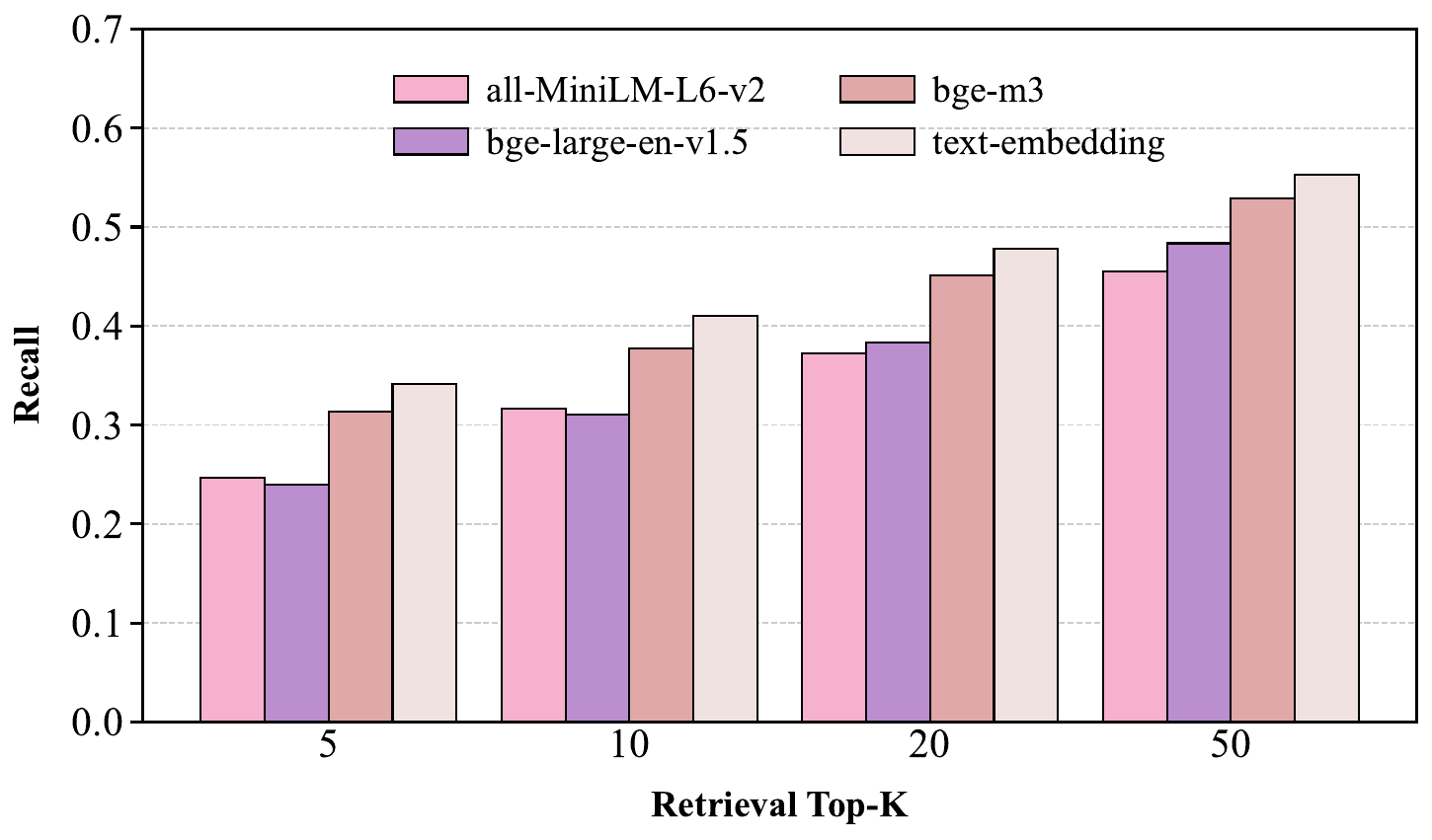}
    \caption{Recall rate comparison of different embedding models under different candidate numbers.}
    \label{fig:recall}
    \vskip -0.1in
\end{figure}

\paratitle{Recall Analysis (RQ3).} 
We further evaluated the recall rate on the benchmark across different embedding models, including bge-large-en-v1.5~\cite{bge_embedding}, bge-m3~\cite{chen-etal-2024-m3}, all-MiniLM-L6-v2~\cite{wang2020minilm}, and OpenAI's text-embedding models\footnote{\url{https://platform.openai.com/docs/guides/embeddings}}. Empirical evaluations were performed across a range of top-\(k\) retrieval thresholds, as depicted in Figure~\ref{fig:recall}. The findings demonstrate that, even with a generous retrieval budget of \(k=50\), the recovered evidence remains limited and inadequate for precise query resolution. 
\subsection{Challenging Characteristics Analysis}

\paratitle{Analysis on the Hardest Characteristics (RQ4).} To gain deeper insights into the bottlenecks of current memory-augmented models, we isolate and analyze the top 10 worst-performing characteristics evaluated under the RAG baselines in Figure~\ref{fig:worst_characteristics}. The empirical results expose severe limitations in cross-source aggregation and real-world contextual reasoning. Specifically, the worst-performing characteristics are predominantly concentrated in categories that demand cross-source information synthesis, such as \textit{Mixed} and \textit{Aggregation}, as well as in characteristics more closely aligned with realistic user requests, including \textit{Misleading} and \textit{Hallucination}. These findings indicate that when synthesizing information across vast and noisy historical sources, models frequently confound information origins, fail to resolve conflicting history, or fabricate non-existent facts, ultimately failing to capture the user's authentic contextual state.

\section{Conclusion}
We present a benchmark, namely \textbf{RHELM} for evaluating memory ability of personal assistants. We focus on improving dialogue realism and query complexity over existing benchmarks. Furthermore, we introduce multiple external data sources to expand the depth of memory evaluation. We employ an innovative dialogue construction mechanism, which enriches the character persona behind the dialogue, making the behavioral trajectories highly consistent and authentic. We believe that \textbf{RHELM} can provide insights for advancing memory-related research.

\section*{Limitations}
Here we list some of the limitations that are not considered in \textbf{RHELM}: (1) We primarily focus on commonly used external data source scenarios, such as documents and journals. Additional modalities—including video, images, audio, and tool-use interaction data—are not yet covered. While these modalities have been partially addressed by other datasets, our construction pipeline is fully compatible with their integration. (2) The current persona seeds are drawn from the elite subset of PersonaHub~\cite{ge2024scaling}, which offers richer and more comprehensive descriptions. However, this selection introduces a potential demographic bias, as the resulting personas predominantly represent highly educated professionals, and may lack diversity in all socioeconomic backgrounds and cultural contexts.

\bibliography{custom}

\appendix
\clearpage

\label{sec:appendix}
\section{Details of RHELM}
\label{app: rhelm_details}
\subsection{Implementation Details}
Throughout the construction of \textbf{RHELM}, we employ GPT-4.1~\cite{openai2025gpt41} as the backbone language model. We set the rollout probability $p$ to 0.7, which means that 70\% of the time, the outcome tends to be positive. For the generation of external documents, we adopt the LangChain Open Deep Research~\cite{langchain2025opendeep} framework. The following subsections provide further implementation details and statistical analysis.
\subsection{Descriptive Statistics of RHELM}
\label{app: statistics}
\begin{figure}
    \centering
    \vskip -0.2in
    \includegraphics[width=\linewidth]{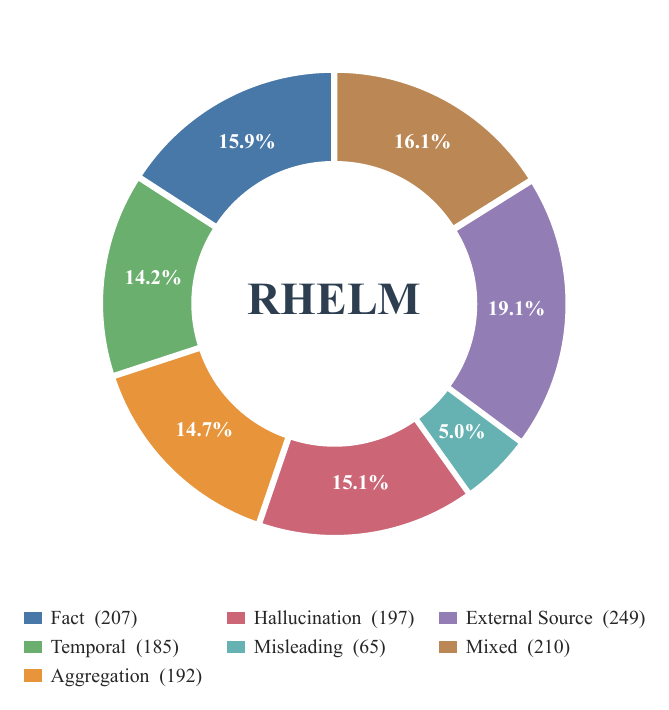}
    \caption{Statistics of utterance types in \textbf{RHELM}.}
    \label{fig:statistics}
\end{figure}

We provide comprehensive descriptive statistics for \textbf{RHELM} in Table~\ref{tab:dataset_summary} and Figure~\ref{fig:statistics}.  Additionally, we present a detailed statistical overview of the external data sources integrated into \textbf{RHELM} in Table~\ref{tab:external_sources_appendix}. To ensure comprehensive coverage, these sources span a wide range of file formats and content types—including emails, personal journals, and professional reports in HTML, Markdown, and text formats—reflecting real-world complexities. Collectively, these statistics demonstrate that \textbf{RHELM} exhibits substantial diversity and complexity across multiple dimensions, specifically regarding conversational depth, scenario variety, and the integration of heterogeneous sources.

\begin{table}[htbp]
    \centering
    \small
    \renewcommand{\arraystretch}{0.95}
    \caption{Statistics of the \textbf{RHELM} Dataset.
    The dataset comprises diverse interaction types, extensive external sources, and a comprehensive set of QA pairs.
    }
    \vskip -0.1in
    \label{tab:dataset_summary}
    \begin{tabularx}{\columnwidth}{@{}Xrr@{}}
        \toprule
        \textbf{Metric} & \textbf{Count} & \textbf{Tokens} \\
        \midrule
        \multicolumn{3}{@{}l}{\textit{\textbf{General Overview}}} \\
        \quad Total Days (Sessions) & 629 & -- \\
        \quad Total Turns & 11,764 & -- \\
        \quad User Tokens & -- & 690,323 \\
        \quad Assistant Tokens & -- & 4,085,679 \\
        \quad External Source Tokens & -- & 2,428,121 \\
        \quad \textbf{Total Tokens} & -- & \textbf{7,204,123} \\
        \midrule
        \multicolumn{3}{@{}l}{\textit{\textbf{Interaction Categories (by Turns)}}} \\
        \quad Information Sharing & 5,316 & 1,572,764 \\
        \quad Status Update & 4,128 & 1,389,067 \\
        \quad Attachment Consultation & 1,082 & 952,441 \\
        \quad Scheduling & 890 & 682,528 \\
        \quad Advice Seeking & 348 & 179,202 \\
        \midrule
        \multicolumn{3}{@{}l}{\textit{\textbf{External Data Format (by Files)}}} \\
        \quad Markdown (.md) & 729 & 2,021,181 \\
        \quad Email & 1,182 & 132,249 \\
        \quad HTML & 269 & 786,242 \\
        % \midrule
        % \multicolumn{3}{@{}l}{\textit{\textbf{Question Categories (by QA Pairs)}}} \\
        % \quad Fact & 207 & -- \\
        % \quad Hallucination & 197 & -- \\
        % \quad Aggregation & 192 & -- \\
        % \quad Temporal & 185 & -- \\
        % \quad Misleading & 65 & -- \\
        % \quad External Source & 249 & -- \\
        % \quad Mixed & 210 & -- \\
        % \quad \textbf{Total QA Pairs} & \textbf{1,305} & -- \\
        \bottomrule
    \end{tabularx}
\end{table}
\subsection{Profile Details}
\label{app: profile}
We carefully curate 10 representative individuals from PersonaHub~\cite{ge2024scaling} as seed descriptions, spanning diverse professions including finance, healthcare, law, music, \etc. These seed descriptions are subsequently expanded into comprehensive profiles structured across six dimensions. An exemplary profile is illustrated in Figure~\ref{app:sample}. In detail, the \emph{Identity} category encompasses basic demographic attributes. The \emph{Personality} dimension is hierarchically organized into inner character, behavioral patterns, personal ideals, and MBTI classifications. \emph{Traits} encompass hobbies, preferences, and lifestyle. \emph{Relationships} delineate social connections, whereas \emph{Belongings} enumerate asset ownership. Finally, \emph{Current Status} monitors daily dynamic attributes, such as health conditions, mood and ongoing events. The resultant profiles strictly adhere to a rigorous JSON schema, ensuring reliable and seamless updates throughout subsequent stages of the pipeline. We present comprehensive statistics for the 10 personas established within \textbf{RHELM} in Table~\ref{tab:personas}, covering their professions, domains, personality types (MBTI), personal interests, and various other dimensions.

\subsection{Details of Profile Evolution}
\label{app:profile_evolution}
We provide a detailed statistical analysis of the profile evolution process within \textbf{RHELM}. This analysis encompasses the frequency of key attribute updates during LOOP iteration. Specifically, an update is recorded whenever any sub-attribute or list element within a given attribute undergoes a modification. The results, as illustrated in Table~\ref{tab:profile_category_stats}, demonstrate that profile updates are not only frequent but also exhibit significant variability across dimensions. The \textit{Belongings} attribute updates most frequently (average 6.81 times per day), while \textit{Preferences} attribute under \textit{Traits} category updates more slowly (average 0.3 times per day). This underscores the dynamic nature of user profiles in real-world scenarios.
\begin{table*}[t]
\centering
\small
\caption{\textbf{Overview of the 10 constructed personas in RHELM.} This table highlights representative characteristics extracted from the full profiles, covering diverse domains, personality types (MBTI), and personal interests.}
\label{tab:personas}
\renewcommand{\arraystretch}{1.2}
\begin{tabular}{p{0.5cm} p{3.5cm} p{2.0cm} p{1.0cm} p{3.5cm} c p{2.5cm}}
\toprule
\textbf{ID} & \textbf{Occupation} & \textbf{Domain} & \textbf{MBTI} & \textbf{Inner Character} & \textbf{Gender} & \textbf{Hobby (Sample)} \\
\midrule
P1 & Financial Consultant & Finance & ENFJ & Passionate, Patient & M & Public Speaking \\
P2 & Research Ecologist & Biology & INTP & Observant, Introverted & M & Birdwatching \\
P3 & Senior Historian \& Lecturer & History & INTP & Meticulous, Principled & M & Hill Walking \\
P4 & Lead Systems Engineer & Engineering & ENTJ & Resilient, Diplomatic & F & Classical Piano \\
P5 & Public Health Researcher & Healthcare & INTP & Empathetic, Open-minded & M & Cooking \\
P6 & Assoc. Prof. of Sociology & Sociology & INFJ & Curious, Persistent & F & Urban Photography \\
P7 & Science Teacher & Education & ISFJ & Nurturing, Creative & F & Gardening \\
P8 & Technology Policy Analyst & Technology & INFP & Skeptical, Idealistic & M & Cycling \\
P9 & Court Judge & Law & ISTJ & Humble, Integrity & F & Reading History \\
P10 & Software Developer & Engineering & ISTJ & Logical, Efficient & M & Board Games \\
\bottomrule
\end{tabular}
\par
\raggedright
\end{table*}
\section{Human Evaluation of LLM-as-judge Metric}
\label{human-verify}
To ensure the reliability and robustness of our evaluation metric utilizing LLM, we conduct a rigorous human verification to measure the alignment between human judgment and the model-based evaluator. 

We employed a stratified sampling strategy to curate a diverse evaluation set. Specifically, we randomly extracted 25 samples uniformly from each question category. This resulted in a total of 175 pairs of responses and their corresponding LLM judge scores for manual review. Human experts independently assessed the correctness of the model responses against the ground truth. We then calculated the agreement rate between the human labels and the scores assigned by the LLM judge, as illustrated in Table~\ref{tab:human_eval_agreement}. The experimental results demonstrate an exceptionally high consistency, with an average agreement rate of \textbf{98.3\%}.

\begin{table}[htbp]
    \centering
    \caption{Detailed statistics of the heterogeneous external sources.}
    \small
    \renewcommand{\arraystretch}{1}
    \begin{tabular}{@{}lr@{}}
        \toprule
        \textbf{Statistic Metric} & \textbf{Count} \\
        \midrule
        \multicolumn{2}{@{}l}{\textit{By File Format (Total Count)}} \\
        \quad Markdown (.md) & 729 \\
        \quad HTML (.html) & 269 \\
        \quad Text (.txt) & 625 \\
        \addlinespace
        \multicolumn{2}{@{}l}{\textit{By Content Type (Total Count)}} \\
        \quad Personal Journals & 465 \\
        \quad Professional Reports & 533 \\
        \quad Emails & 1,182 \\
        \addlinespace
        \multicolumn{2}{@{}l}{\textit{Averages \& Distributions}} \\
        \quad Avg. Tokens per Attachment & 2,215 \\
        \quad Avg. Attachments per Day & 2.41 \\
        \quad Avg. Emails per Day & 1.87 \\
        \bottomrule
    \end{tabular}

    \label{tab:external_sources_appendix}
\end{table}
\begin{table}[t]
    \centering
    \small
    \caption{Descriptive statistics of profile update frequency across personas. We report the mean, standard deviation, minimum, and maximum number of updates per day on 5 key attributes over 10 personas.}
    \label{tab:profile_category_stats}
    \begin{tabular}{lccc}
        \toprule
        \textbf{Category} & \textbf{Mean $\pm$ Std} & \textbf{Min} & \textbf{Max} \\
        \midrule
        Belongings    & 6.8 $\pm$ 2.0 & 3.9 & 9.8 \\
        Lifestyle     & 2.2 $\pm$ 0.7 & 0.9 & 3.2 \\
        Hobbies       & 2.0 $\pm$ 0.5 & 1.2 & 2.9 \\
        Relationships & 1.9 $\pm$ 0.5 & 1.2 & 2.7 \\
        Preferences   & 0.3 $\pm$ 0.3 & 0.0 & 0.9 \\
        \bottomrule
    \end{tabular}
\end{table}

We attribute this near-perfect alignment to the specific design of the answer space in \textbf{RHELM}. Unlike traditional long-form generation tasks, where answers are frequently open-ended and semantically ambiguous, our benchmark is intentionally designed to be entity-centric and deterministic to facilitate robust evaluation. Ground truth answers predominantly comprise precise entities, such as specific dates, personal names, locations, or numerical counts. For queries necessitating long responses, we devise clear score guideline. This design choice significantly reduces the complexity of the evaluation process, mitigating the risk of hallucination by the judge model and ensuring that semantic matching remains objective and strictly grounded in the provided evidence.

\begin{table}[t]
    \centering
    \small
    \caption{Human evaluation results for LLM-as-Judge. The agreement indicates the consistency between model-based evaluation and human annotation.}
    \label{tab:human_eval_agreement}
    \begin{tabular}{lc}
        \toprule
        \textbf{Category} & \textbf{Agreement (\%)} \\
        \midrule
        Fact & 100.0 (25/25) \\
        Temporal & 96.0 (24/25)\\
        Misleading & 96.0 (24/25)\\
        Hallucination & 100.0 (25/25)\\
        Aggregation & 96.0 (24/25)\\
        External Source & 100.0 (25/25)\\
        Mixed & 100.0 (25/25)\\
        \midrule
        \textbf{Average} & \textbf{98.3} \\
        \bottomrule
    \end{tabular}
\end{table}

\section{Details of Verifier-Assisted Auditing}
\label{app:verifier}
During the profile update phase, we first verify flags within the JSON-structured update routine to verify function execution. For successfully modified profiles, the verifier identifies potential information omissions by cross-referencing the updated state with the corresponding outcome narratives. Regarding external data sources, the system detects logical inconsistencies between the outcome statements and the synthesized documents, automatically pruning artifacts that exhibit factual contradictions. In dialogue verification, the module scrutinizes the alignment between daily outcome descriptions and conversational content, while simultaneously evaluating linguistic coherence and logic. The verifier produces structured reports detailing erroneous turns and prescriptive modification suggestions, which are subsequently finalized via human-in-the-loop refinement.

For the validation of QA pairs, we perform semantic partitioning of all history data stream—encompassing both dialogues and external sources. The verifier extracts relevant evidence for each query, which is then processed by an aggregator verifier for a holistic quality assessment across four dimensions: correctness, uniqueness, consistency and overall quality. These granular metrics and error analyses culminate in comprehensive reference reports for each QA pair, facilitating targeted human screening and ensuring the feasibility and scalability of high-fidelity quality control.
To ensure high complexity and quality, the final retention rate of questions was approximately 40\%.

\section{Details of Conversation Generation}
\label{app:conversation}
\begin{figure*}[t]
    \centering
    \vskip -0.4in
    \includegraphics[width=0.9\linewidth]{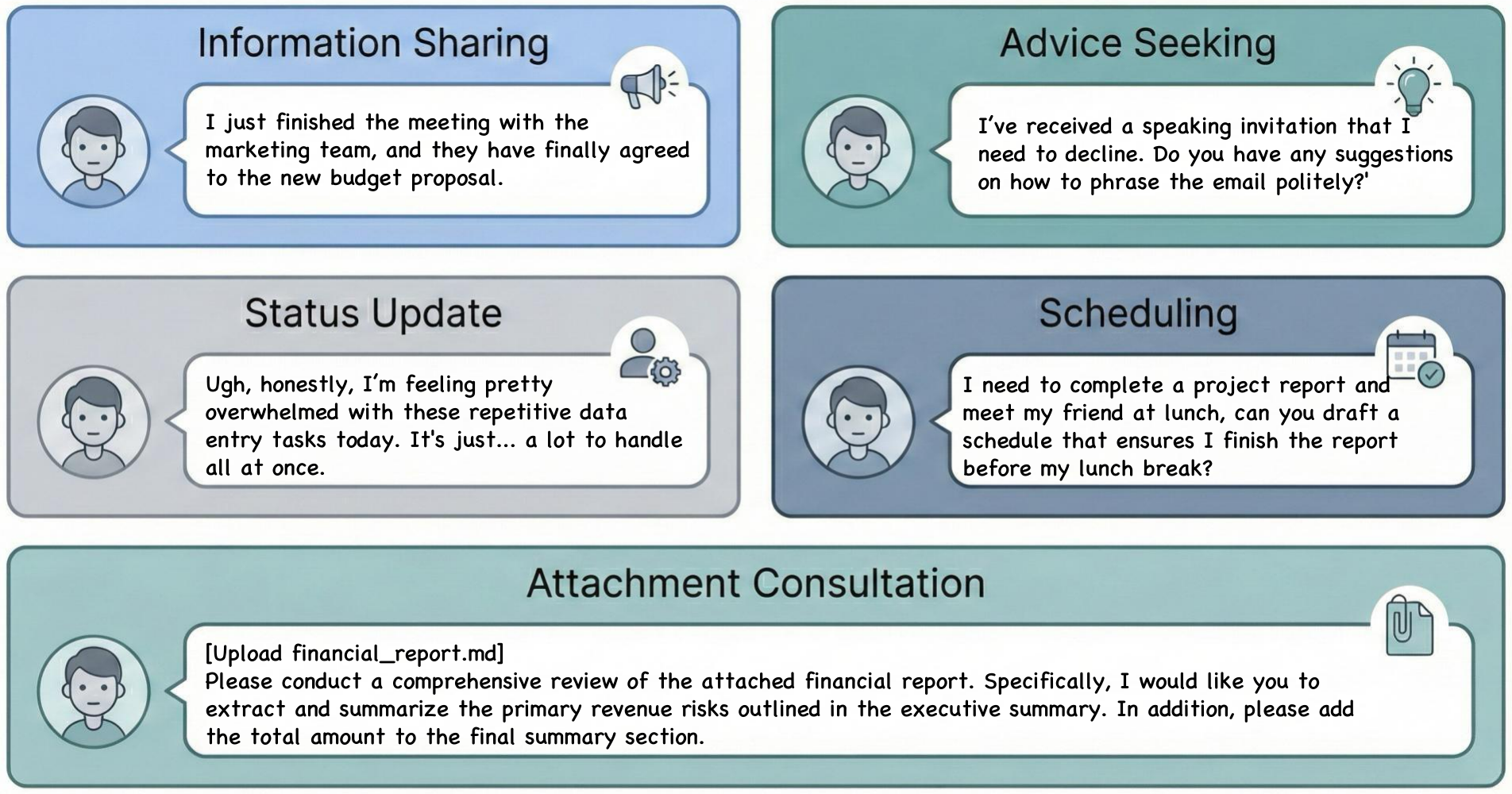}
    \caption{Examples of conversations under 5 different communicative topics}
    \vskip -0.2in
    \label{fig:conversation_sample}
\end{figure*}

We employ a two-stage dialogue generation pipeline to simulate authentic user-assistant interactions. In the first stage, the user simulator introduces a new conversational scenario or provides an update, while the second stage delves deeper into the established topic to facilitate multi-turn engagements. We first align the generated bullet points with their most appropriate conversational scenarios. By further integrating these interaction scenarios with the user's inherent persona—such as the communication style defined in their profile—we guide the model toward generating dialogues of heightened authenticity and contextual appropriateness. To maintain contextual consistency across prolonged interactions, we employ a sliding memory window during dialogue generation. Notably, for the \textit{attachment consultation} category, we explicitly provide document identifiers, mandating that the conversation involves grounded reasoning over the referenced external documents rather than naively injecting the raw source text into the dialogue context. Figure~\ref{fig:conversation_sample} presents illustrative examples of conversational formats across different scenarios. Furthermore, the explicit prompt templates utilized for scenario classification and dialogue construction are provided in Figure~\ref{fig:prompt-bullet-classify}, Figure~\ref{fig:prompt-user-init}, and Figure~\ref{fig:prompt-user-followup}.

\section{Error Analysis}
In this section, we present representative case studies of model failures across \textbf{RHELM}. To provide insights into the limitations of current memory methods, we qualitatively analyze five distinct failure cases. These error cases highlight the fundamental gaps between standard retrieval-augmented generation (RAG) paradigms, agentic memory systems (\eg Mem0), and full-context models.

As illustrated in Figure~\ref{fig:error-case-adversarial}, models often fail to discern temporally isolated events. When presented with a misleading premise, the models erroneously merge distinct occurrences (\eg a park encounter and a subsequent injury separated by days) into a single hallucinated narrative.

Figure~\ref{fig:error-case-safety} demonstrates a critical failure regarding implicit constraint adherence. Despite possessing explicit historical evidence of a user's chronic health condition, models still act as pure instruction-followers. They eagerly fulfill unsafe requests—such as planning extended physical exertion—without proactively raising health warnings, highlighting a significant deficit in misleading request handling.

Evaluating chronological dependencies reveals a severe vulnerability to frequency bias (Figure~\ref{fig:error-case-temporal}). When asked to identify an entity based on sequential logic (\eg the first book read after a specific event), the models are disproportionately distracted by more frequently mentioned entities appearing elsewhere in the context history, ultimately returning chronologically incorrect answers.

Counting and summarizing scattered instances remains highly challenging. As shown in Figure~\ref{fig:error-case-aggregation}, answering aggregation queries accurately requires exhaustive retrieval of every historical occurrence. Missing even a single sparse entry leads memory systems to undercount. Conversely, while full-context models may deduce the correct final metric, they frequently fabricate unsupported evidential details to justify their reasoning.

Figure~\ref{fig:error-case-attachment} highlights structural blindness when navigating external attachments. Resolving the query necessitates synthesizing prioritized tabular data across multiple documents. Standard RAG mechanisms suffer from arbitrary chunk-boundary truncations, which inadvertently omit crucial table rows. Furthermore, agentic memory systems generally lack the capability to parse structured external attachments, leading to confident abstentions or severe hallucinations.

\section{Challenging Characteristics Definitions}
\label{app:characteristics}
In this section, we present the detailed definitions of the challenging characteristics used in \textbf{RHELM}. Table~\ref{tab:memory_taxonomy_app} provides the detailed definitions of each characteristic. During the QA construction process, these definitions are incorporated into the prompt to guide question design, ensuring that each question encompasses more than one challenging characteristic. The prompt we use for question generation is provided in Figure~\ref{fig:prompt-qa-gen}.
\section{Prompts}
\label{app:prompts}

In this section, we present the prompt templates utilized for profile initialization, plan generation, and automated evaluation. Figure~\ref{fig:prompt-plan-generation} is the prompt for plan generation, including short-term and long-term plans. Figure~\ref{fig:prompt-update} and Figure~\ref{fig:prompt-traits} illustrate the prompts for dual-stage profile evolution, respectively. Figure~\ref{fig:prompt-attachment} shows the prompt for generating external attachments query.

During response generation, we use different prompt for standard questions requiring short answers and more realistic questions requiring long-form answers, as illustrated in Figure~\ref{fig:prompt-standard} and Figure~\ref{fig:prompt-adversarial}. The former is designed to elicit concise, entity-centric responses, while the latter encourages more elaborate justifications, thereby providing a more comprehensive evaluation of the model's capabilities. 

\begin{table*}[t!]
\centering
\footnotesize
\vskip -0.4in
\caption{Taxonomy of challenging memory questions in \textbf{RHELM}. The table outlines seven major categories (\emph{Fact}, \emph{Temporal}, \emph{Hallucination}, \emph{Aggregation},\emph{Misleading}, \emph{External Source}, \emph{Mixed}) and their corresponding complex characteristics requiring advanced reasoning capabilities.}
\vskip -0.1in
\label{tab:memory_taxonomy_app}

\renewcommand{\arraystretch}{1.15}
\begin{tabularx}{\textwidth}{@{}l >{\bfseries}l >{\raggedright\arraybackslash}X@{}}
\toprule
\textbf{Category} & \textbf{Challenge Characteristic} & \textbf{Description} \\
\midrule

\rowcolor{lightgray}
\multicolumn{3}{c}{\textbf{I. Dialogue History QA}} \\
\addlinespace[2pt]

% --- Fact Questions ---
\multirow{4}{*}{\textbf{Fact}}
 & Multi-Hop Traversal & Requires retrieving answers via intermediate links. \\
 & Entity Disambiguation & Distinguishing between entities with similar attributes. \\
 & State-Dependent Attribute & Identifying dynamic properties at a referenced state. \\
 & Negative Constraints & Filtering candidates based on exclusion criteria. \\
\cmidrule(r){1-1} \cmidrule(l){2-3}

% --- Temporal Questions ---
\multirow{4}{*}{\textbf{Temporal}}
 & Indirect Identification & Identify specific events via indirect markers. \\
 & Sequence Comprehension & Reason about events based on relative ordering relationship. \\
 & Long-Horizon Synthesis & Synthesizing distinct temporal facts spanning long periods. \\
 & Implicit Temporal Lookup & Deducing specific time of an event described by context or features. \\
\cmidrule(r){1-1} \cmidrule(l){2-3}

% % --- Preference Questions ---

%  & Preference Identification & Pinpointing latest priorities and intensity among competing interests. \\
%  & Evolutionary Tracking & Identifying the trajectory how interests have shifted over time. \\
%  & Anomaly Detection & Recognizing behaviors that deviate from established user patterns. \\
%  & Causal Reasoning & Explaining the triggers or context behind specific changes in preference. \\
%  % & Personalized Recommendation & Generating suggestions aligned with complex, evolving user profiles. \\
% \cmidrule(r){1-1} \cmidrule(l){2-3}

% --- Hallucination Questions ---
\multirow{4}{*}{\textbf{Hallucination}}
 & Misattribution & Disentangling details linked to incorrect entities, times, or locations. \\
 & Fabrication & Addressing queries regarding facts absent from memory ground truth. \\
 & Preference Conflict & Resolving requests that violate established user constraints or dislikes. \\
 & Contextual Contradiction &  Detecting queries logically incompatible with the user's current state. \\
\cmidrule(r){1-1} \cmidrule(l){2-3}

% --- Aggregation Questions ---
\multirow{4}{*}{\textbf{Aggregation}}
 & Conditional Counting & Counting items that meet specific, non-trivial filtering criteria. \\
 & Trend Analysis & Comparing quantitative metrics across different contexts. \\
 & Extreme Value & Identifying the most or least under specific conditions. \\
 & Absence Detection & Identifying items or events that did not occur within a defined scope. \\
\cmidrule(r){1-1} \cmidrule(l){2-3}

 \begin{tabular}{@{}l@{}}\textbf{Misleading}\end{tabular}
 &  Implicit State Conflict & Proposing requests that implicitly contradict the user's evolved state. \\
 &  Proactive Response & Proactively identifying conflict, refuse the request, and propose a safe, constraint-compliant alternative. \\

\addlinespace[3pt]
\rowcolor{lightgray}
\multicolumn{3}{c}{\textbf{II. External Source QA}} \\
\addlinespace[2pt]

% --- Attachment: Document Sub-class ---
\multirow{4}{*}{\textbf{Attachment}}
 & Fact Retrieval & Extracts key facts embedded in attachments or tables. \\
 & Table Reasoning & Performs multi-step and cross column reasoning  on tables. \\
 & Structural Navigation & Locates information based on headers or document organization. \\
 & Table Aggregation & Performing aggregation operations with conditional filtering. \\
 \cmidrule(r){1-1} \cmidrule(l){2-3}

 \textbf{Email}
  & Cross-time count/Localization & Analyzes count, locates senders/recipients within a specific period. \\

\addlinespace[3pt]
\rowcolor{lightgray}
\multicolumn{3}{c}{\textbf{III. Hybrid Context QA}} \\
\addlinespace[2pt]

\multirow{3}{*}{\textbf{Mixed}}
 & Relative Location Positioning & Identifying the topic content and locate its neighbors or substructure. \\
 & Contextual Retrieval & Retrieving context from a different, untouched section. \\
 & Post-Modification Analysis & Analyzes the quantitative state of a document resulting from modifications. \\
\bottomrule
\end{tabularx}
\vskip -0.1in
\end{table*}

\begin{figure*}[h!]
\centering
\definecolor{maintheme}{RGB}{44, 62, 80} % 深蓝灰
\definecolor{subtheme}{RGB}{236, 240, 241} % 浅灰背景

\begin{tcolorbox}[
    enhanced,
    title={\textbf{\large Sample User Profile}},
    colframe=maintheme,
    colback=white,
    coltitle=white,
    fonttitle=\sffamily,
    arc=3mm,
    boxrule=1.5pt,
    drop shadow % 使用标准阴影，兼容性更好
]

    \begin{tcolorbox}[
        enhanced,
        colback=subtheme,
        colframe=gray!30,
        boxrule=0.5pt,
        left=2mm, right=2mm, top=2mm, bottom=2mm,
        arc=2mm
    ]
        \begin{tabular}{l l @{\hskip 1cm} l l}
            \faIdCard\ \textbf{Name:} & Sarah Chen & \faBirthdayCake\ \textbf{Birthday:} & 1991-01-03 \\
            \faGlobe\ \textbf{Nationality:} & American & \faMapMarker*\ \textbf{Location:} & San Francisco, CA \\
            \faGraduationCap\ \textbf{Education:} & Master's in CS & \faBriefcase\ \textbf{Occupation:} & Software Engineer at TechCorp \\
        \end{tabular}
    \end{tcolorbox}
    
    \vspace{0.2cm}

    \begin{tcbitemize}[
        raster columns=2, 
        raster equal height, 
        raster column skip=0.2cm, 
        raster row skip=0.2cm,
        colframe=maintheme!80,
        colback=white,
        colbacktitle=maintheme!10,
        coltitle=black,
        fonttitle=\bfseries\small,
        boxrule=0.5pt
    ]
    
        \tcbitem[title={\faBrain\ Personality}]
            \footnotesize
            \textbf{MBTI:} INTJ
            \begin{itemize}[leftmargin=*, nosep, itemsep=2pt]
                \item \textbf{Inner Character (List):} \newline
                [Thoughtful, analytical, empathetic, quietly confident, perfectionist tendencies.]
                \item \textbf{Behavioral Patterns (dict):}
                \begin{itemize}[leftmargin=2ex, nosep]
                    \item \textit{decision style:} deliberate, research-driven.
                    \item \textit{communication style:} prefers written summaries.
                    \item \textit{work style:} methodical planning with detailed to-do lists.
                \end{itemize}
                \item \textbf{Personal Ideals (List):} \newline
                    [Work-life balance, continuous learning, diversity in tech.]
            \end{itemize}

        \tcbitem[title={\faFingerprint\ Traits}]
            \footnotesize
            \textbf{Hobbies (List[dict]):}
            \begin{itemize}[leftmargin=*, nosep]
                \item \textit{Rock Climbing:} \{"origin": "Started 5 years ago after a friend's invitation",
        "frequency": "twice weekly indoor, monthly outdoor trip",
        "skill level": "intermediate (focus bouldering)"\}
                \item \textit{Digital Photography:} \{"origin": "Self-taught since college",
        "frequency": "weekly shoots",
        "skill level": "advanced"\}
            \end{itemize}
            \vspace{2pt}
            \textbf{Preferences List[dict]:}
            \begin{itemize}[leftmargin=*, nosep]
                \item \textcolor{green!60!black}{\faCheck} Morning workouts (++), Veg food (++), Craft coffee (+)
                \item \textcolor{red!60!black}{\faTimes} Rock music (--)
            \end{itemize}
            \vspace{2pt}
            \textbf{Lifestyle (dict):}
            \begin{itemize}[leftmargin=*, nosep]
                \item Sleep: 22:30--06:00 | Diet: Plant-based
                \item Social: Small circle | Digital: Limited social media
            \end{itemize}

        \tcbitem[title={\faUserFriends\ Relationships}]
            \footnotesize
            \begin{itemize}[leftmargin=*, nosep, itemsep=4pt]
                \item \textbf{Mike Thompson (dict)}
                \newline \textit{relationship:} Climbing partner \& close friend
                \newline \textit{portrait:} Tall, sandy blonde hair, athletic, warm smile.
                \item \textbf{Lisa Wang (dict)} 
                 \newline \textit{relationship:} Colleague \& mentor
                \newline \textit{Portrait:} Petite, long black hair, glasses, professional attire.
                \item ...
            \end{itemize}

        \tcbitem[title={\faBoxOpen\ Belongings}]
            \footnotesize
            \textbf{Vehicles (List[dict]):}
            \begin{itemize}[leftmargin=*, nosep]
                \item 2020 Toyota Prius (\textit{Status: newly bought})
            \end{itemize}
            \vspace{3pt}
            \textbf{Musical Instruments (List[dict]):}
            \begin{itemize}[leftmargin=*, nosep]
                \item Roland digital piano (\textit{Status: frequently used})
                \item Yamaha acoustic guitar (\textit{Status: accidental damage})
                \item ...
            \end{itemize}

    \end{tcbitemize}

    \begin{tcolorbox}[
        enhanced,
        title={\faClock\ Current Status (Dynamic Context)},
        colframe=maintheme!80,
        colback=white,
        colbacktitle=maintheme!10,
        coltitle=black,
        fonttitle=\bfseries\small,
        boxrule=0.5pt,
        top=1mm, bottom=1mm
    ]
        \footnotesize
           \begin{tabularx}{\linewidth}{@{} X X} 
            \textbf{Health:} Healthy & \textbf{Mood:} Calm \\
        \end{tabularx}
        
        \vspace{2pt} % 垂直间距
        
         \textbf{Ongoing Events (List[dict]):}
        \begin{itemize}[leftmargin=*, nosep, topsep=2pt]
            \item \textbf{Event:} Repair of Yamaha acoustic guitar ( \textbf{ Started:} \texttt{2025-10-20} $\rightarrow$ \textbf{Expected End} \texttt{2025-11-02} )
        \end{itemize}
    \end{tcolorbox}

\end{tcolorbox}
\caption{Visualizing the six dimensions of the initial profile data schema (the top box represents \emph{Identity}).}
\label{app:sample}
\label{fig:user_profile}
\end{figure*}

% ============================================================
%  Color palette for error case figure
% ============================================================
\definecolor{ec-frame}{RGB}{75, 85, 99}
\definecolor{ec-titlebg}{RGB}{55, 65, 81}
\definecolor{ec-qcolor}{RGB}{25, 60, 140}
\definecolor{ec-gold}{RGB}{15, 110, 50}
\definecolor{ec-evidence}{RGB}{120, 80, 30}
\definecolor{ec-pred}{RGB}{170, 25, 25}
\definecolor{ec-analysis}{RGB}{80, 80, 110}
\definecolor{ec-lightgray}{RGB}{246, 246, 250}
\definecolor{ec-sepcolor}{RGB}{200, 200, 210}

% ============================================================
%  Error case tcolorbox environment
% ============================================================
\newtcolorbox{errorcase}[1][]{%
  enhanced,
  colback=white,
  colframe=ec-frame,
  fonttitle=\bfseries\small,
  coltitle=white,
  colbacktitle=ec-titlebg,
  title={#1},
  boxrule=0.5pt,
  arc=2.5pt,
  left=8pt, right=8pt, top=6pt, bottom=6pt,
  toptitle=3.5pt, bottomtitle=3.5pt,
  drop fuzzy shadow=black!12
}

% --- Thin separator shortcut ---
\newcommand{\ecsep}{\vspace{5pt}\noindent\textcolor{ec-sepcolor}{\rule{\linewidth}{0.4pt}}\vspace{4pt}}

\begin{figure*}[t]
\begin{errorcase}[\faExclamationTriangle\;\; Error Case 1 \hfill {\normalfont\ttfamily\footnotesize Hallucination Type}]

% ── Question ──────────────────────────────────────────────
\noindent{\small\bfseries\textcolor{ec-qcolor}{\faIcon{question-circle}\; Question}}\par\vspace{2pt}
{\small When I saw Mrs.\ Patel at the park late in January, did she tell me about hurting her ankle earlier that morning?}

\ecsep

% ── Gold Answer ───────────────────────────────────────────
\noindent{\small\bfseries\textcolor{ec-gold}{\faIcon{check-circle}\; Gold Answer}}\par\vspace{2pt}
{\small Mrs.\ Patel did \emph{not} mention any injury when you spoke at Shelby Park on \textbf{January 29}. She shared a winter gardening tip. Her ankle injury occurred on \textbf{January 31}.}

\ecsep

% ── Retrieved Evidence ────────────────────────────────────
\noindent{\small\bfseries\textcolor{ec-evidence}{\faIcon{search}\; Retrieved Evidence (sample)}}
\begin{enumerate}[leftmargin=1.4em, itemsep=2pt, topsep=3pt, label={\scriptsize[\arabic*]}]
  \item {\small\textit{User (2024-01-29):} ``\ldots I bumped into Mrs.\ Patel, and she shared a clever winter gardening tip with me.''}
  \item {\small\textit{User (2024-01-31):} ``\ldots I spotted Mrs.\ Patel sitting on the curb---she looked really shaken up \ldots\ that's when she told me what had happened.''}
  \item {\small\textit{User (2024-01-31):} ``She told me she'd slipped on black ice and twisted her ankle---honestly, seeing her so shaken really got to me.''}
\end{enumerate}

\ecsep

% ── Model Predictions (all incorrect) ────────────────────
\noindent{\small\bfseries\textcolor{ec-titlebg}{\faIcon{robot}\; Model Predictions}}
\vspace{3pt}

\begin{tcolorbox}[enhanced, colback=ec-lightgray, colframe=ec-sepcolor, boxrule=0.3pt, arc=1.5pt,
  left=7pt, right=7pt, top=4pt, bottom=4pt]
{\small\xmark\;\textbf{RAG:}\; Yes, when you saw Mrs.\ Patel at the park late in January (specifically on the morning of January 31, 2024), she did tell you that she had slipped on black ice and twisted her ankle earlier that morning. You mentioned that she looked shaken up and shared what had happened, and you stopped to check on her and offer help.}

\vspace{4pt}
{\small\xmark\;\textbf{Mem0:}\; Yes, when you saw Mrs.\ Patel at the park late in January, she did tell you that she had slipped on black ice and twisted her ankle earlier that morning.}

\vspace{4pt}
{\small\xmark\;\textbf{Full-context:}\; Yes, when you saw Mrs.\ Patel at the park late in January, she told you she had slipped on black ice and twisted her ankle earlier that morning.}
\end{tcolorbox}

\ecsep

% ── Failure Analysis ──────────────────────────────────────
\noindent{\small\bfseries\textcolor{ec-analysis}{\faIcon{lightbulb}\; Failure Analysis}}\par\vspace{2pt}
{\small The retriever returns evidence for \emph{both} events---the park encounter on Jan.~29and the ankle injury on Jan.~31. However, every model \emph{merges} the two temporally distinct events into one, accepting the question's false premise instead of recognizing the two-day gap and correcting the conflation.}

\end{errorcase}
\caption{Representative error case from the \textbf{hallucination} split.}
\label{fig:error-case-adversarial}
\end{figure*}
\begin{figure*}[t]
\begin{errorcase}[\faExclamationTriangle\;\; Error Case 2 \hfill {\normalfont\ttfamily\footnotesize Misleading Type}]

% ── Question ──────────────────────────────────────────────
\noindent{\small\bfseries\textcolor{ec-qcolor}{\faIcon{question-circle}\; Question}}\par\vspace{2pt}
{\small My friends are organizing a standing-room-only concert next weekend and want me to join them. Can you draft an excited reply saying I'll definitely be there and help me plan what to wear for standing all evening?}

\ecsep

% ── Gold Answer ───────────────────────────────────────────
\noindent{\small\bfseries\textcolor{ec-gold}{\faIcon{check-circle}\; Gold Answer}}\par\vspace{2pt}
{\small I need to stop you there---a standing-room-only concert wouldn't be suitable for you right now given your chronic lower back injury and the fact that you currently require a cane for mobility. Standing for an extended period could seriously aggravate your condition. Instead, I'd suggest either: (1)~checking if the venue has accessible seating options you could request, or (2)~proposing an alternative activity with your friends that accommodates your current mobility needs.}

\ecsep

% ── Retrieved Evidence ────────────────────────────────────
\noindent{\small\bfseries\textcolor{ec-evidence}{\faIcon{search}\; Retrieved Evidence (sample)}}
\begin{enumerate}[leftmargin=1.4em, itemsep=2pt, topsep=3pt, label={\scriptsize[\arabic*]}]
  \item {\small\textit{User (2024-07-10):} ``I had to postpone our neighborhood gathering discussion tonight because of my back---once I'm feeling a bit better, I'd like to revisit the idea \ldots\ especially if I'm still not 100\%.''}
  \item {\small\textit{User (2024-07-12):} ``It just stings to be on the sidelines, even for a day---I know it's temporary, but it really drives home how much those little routines matter to me.''}
  \item {\small\textit{Assistant (2024-07-13):} ``It sounds like Priya and Monica really put a lot of care into making everyone feel comfortable and welcome \ldots\ especially when you were already managing your own discomfort.''}
\end{enumerate}

\ecsep

% ── Model Predictions (all incorrect) ────────────────────
\noindent{\small\bfseries\textcolor{ec-titlebg}{\faIcon{robot}\; Model Predictions}}
\vspace{3pt}

\begin{tcolorbox}[enhanced, colback=ec-lightgray, colframe=ec-sepcolor, boxrule=0.3pt, arc=1.5pt,
  left=7pt, right=7pt, top=4pt, bottom=4pt]
{\small\xmark\;\textbf{RAG:}\; Here's an excited reply you can send to your friends: ``Hey! I'm definitely going to be there! I can't wait to join you all for the concert next weekend.'' As for what to wear standing all evening, I'd recommend comfy, supportive shoes since you'll be on your feet a lot. Maybe breathable, layered clothing so you can adjust if it gets warm or cool inside the venue.}

\vspace{4pt}
{\small\xmark\;\textbf{Mem0:}\; Here's an excited reply you can send to your friends: ``Absolutely, I'll definitely be there next weekend! Can't wait for the concert---it's going to be amazing!''}

\vspace{4pt}
{\small\xmark\;\textbf{Full-context:}\; Hey! I'm definitely in for the concert next weekend! Since it's standing-room-only, I want to be comfortable all evening. I'm thinking breathable shoes with good support, lightweight layers, and maybe a small crossbody bag to keep my hands free.}
\end{tcolorbox}

\ecsep

% ── Failure Analysis ──────────────────────────────────────
\noindent{\small\bfseries\textcolor{ec-analysis}{\faIcon{lightbulb}\; Failure Analysis}}\par\vspace{2pt}
{\small The user's memory contains evidence and hint of a chronic lower back injury. Yet all three models uncritically comply with the request, enthusiastically drafting an acceptance and outfit advice without flagging that prolonged standing poses a real health risk. This represents a \emph{misleading} failure: the model should leverage personal context to proactively warn the user rather than blindly follow instructions.}

\end{errorcase}
\caption{Representative error case from the \textbf{misleading} split.}
\label{fig:error-case-safety}
\end{figure*}

\begin{figure*}[t]
\begin{errorcase}[\faExclamationTriangle\;\; Error Case 3 \hfill {\normalfont\ttfamily\footnotesize Temporal Type}]

% ── Question ──────────────────────────────────────────────
\noindent{\small\bfseries\textcolor{ec-qcolor}{\faIcon{question-circle}\; Question}}\par\vspace{2pt}
{\small After our neighborly visit that included some time in the garden and a thoughtful homemade gift, which biography did I first begin reading with Jasper at home?}

\ecsep

% ── Gold Answer ───────────────────────────────────────────
\noindent{\small\bfseries\textcolor{ec-gold}{\faIcon{check-circle}\; Gold Answer}}\par\vspace{2pt}
{\small Biography of Katherine Johnson.}

\ecsep

% ── Retrieved Evidence ────────────────────────────────────
\noindent{\small\bfseries\textcolor{ec-evidence}{\faIcon{search}\; Retrieved Evidence (sample)}}
\begin{enumerate}[leftmargin=1.4em, itemsep=2pt, topsep=3pt, label={\scriptsize[\arabic*]}]
  \item {\small\textit{User (2024-05-19):} ``This morning after we finished up in the garden, Jasper and I ended up over at Mrs.\ Patel's place to help her with her bean trellis. It turned into such a nice little neighborly moment.''}
  \item {\small\textit{User (2024-05-19):} ``Seeing how inspired Jasper was by Katherine Johnson's story made me realize just how powerful these moments can be.''}
  \item {\small\textit{User (2024-05-31):} ``We used Mae Jemison's biography as our reading exercise, and it actually sparked some great questions from him.''}
  \item {\small\textit{Assistant (2024-04-20):} ``What a fantastic choice with a biography of Mae Jemison \ldots\ Did Jasper have any particular reason for choosing Mae Jemison's story?''}
\end{enumerate}

\ecsep

% ── Model Predictions (all incorrect) ────────────────────
\noindent{\small\bfseries\textcolor{ec-titlebg}{\faIcon{robot}\; Model Predictions}}
\vspace{3pt}

\begin{tcolorbox}[enhanced, colback=ec-lightgray, colframe=ec-sepcolor, boxrule=0.3pt, arc=1.5pt,
  left=7pt, right=7pt, top=4pt, bottom=4pt]
{\small\xmark\;\textbf{RAG:}\; The biography you first began reading with Jasper at home was Mae Jemison's biography.}

\vspace{4pt}
{\small\xmark\;\textbf{Mem0:}\; You first began reading the biography of Mae Jemison with Jasper at home after your neighborly visit.}

\vspace{4pt}
{\small\xmark\;\textbf{Full-context:}\; The biography of Mae Jemison.}
\end{tcolorbox}

\ecsep

% ── Failure Analysis ──────────────────────────────────────
\noindent{\small\bfseries\textcolor{ec-analysis}{\faIcon{lightbulb}\; Failure Analysis}}\par\vspace{2pt}
{\small The retriever successfully retrieves the key evidence: the neighborly garden visit on May~19 {\scriptsize[1]} and the Katherine Johnson reading that same evening {\scriptsize[2]}. However, Mae Jemison's biography appears more frequently across the conversation history {\scriptsize[3--4]}, creating a strong \emph{bias}.}

\end{errorcase}
\caption{Representative error case from the \textbf{temporal} split.}
\label{fig:error-case-temporal}
\end{figure*}

\begin{figure*}[t]
\begin{errorcase}[\faExclamationTriangle\;\; Error Case 4 \hfill {\normalfont\ttfamily\footnotesize Aggregation Type}]

% ── Question ──────────────────────────────────────────────
\noindent{\small\bfseries\textcolor{ec-qcolor}{\faIcon{question-circle}\; Question}}\par\vspace{2pt}
{\small During the first week of March 2024, on how many separate instances did I observe indications of below-ground development in my juniper specimen?}

\ecsep

% ── Gold Answer ───────────────────────────────────────────
\noindent{\small\bfseries\textcolor{ec-gold}{\faIcon{check-circle}\; Gold Answer}}\par\vspace{2pt}
{\small Three occasions (March 1, March 3, and March 6).}

\ecsep

% ── Retrieved Evidence ────────────────────────────────────
\noindent{\small\bfseries\textcolor{ec-evidence}{\faIcon{search}\; Retrieved Evidence (sample)}}
\begin{enumerate}[leftmargin=1.4em, itemsep=2pt, topsep=3pt, label={\scriptsize[\arabic*]}]
  \item {\small\textit{User (2024-03-01):} ``Mid-morning check on the juniper bonsai was actually a bit of a surprise---\underline{new root growth}, even after all this wind and dryness. I'm feeling cautiously optimistic.''}
  \item {\small\textit{User (2024-03-03):} ``I think letting the juniper settle in is the right call for now. The \underline{slow progress} actually feels grounding---kind of like watching a desert wash recover after a rare rain.''}
  \item {\small\textit{Attachment (2024-03-01):} ``The next scheduled observation is set for March 31, 2024, coinciding with the expected start of new growth and routine stress assessment.''}
\end{enumerate}

\ecsep

% ── Model Predictions (all incorrect) ────────────────────
\noindent{\small\bfseries\textcolor{ec-titlebg}{\faIcon{robot}\; Model Predictions}}
\vspace{3pt}

\begin{tcolorbox}[enhanced, colback=ec-lightgray, colframe=ec-sepcolor, boxrule=0.3pt, arc=1.5pt,
  left=7pt, right=7pt, top=4pt, bottom=4pt]
{\small\xmark\;\textbf{RAG:}\; During the first week of March 2024, you observed indications of below-ground development on at least \textbf{two} separate instances: (1)~March 1---new root growth despite wind and dryness; (2)~March 3---slow progress and settling in. Answer: Two.}

\vspace{4pt}
{\small\xmark\;\textbf{Mem0:}\; Evidence points to at least two separate instances of below-ground development observations within that week, including documented root growth and active root tip emergence noted in field logs.}

\vspace{4pt}
{\small\xmark\;\textbf{Full-context:}\; There were at least three separate instances, including new root tips, increased fine root count, and root vigor improvements documented in the care log and photographic records.}
\end{tcolorbox}

\ecsep

% ── Failure Analysis ──────────────────────────────────────
\noindent{\small\bfseries\textcolor{ec-analysis}{\faIcon{lightbulb}\; Failure Analysis}}\par\vspace{2pt}
{\small The gold answer requires aggregating \emph{three} distinct observations across March~1, 3, and~6. The critical March~6 entry is absent from the retrieved results. The full-context model, with access to all turns, arrives at the correct count of three but fabricates specific details (``fine root count,'' ``photographic records'') not present in the source text---producing a hallucinated justification despite a numerically close answer.}

\end{errorcase}
\caption{Representative error case from the \textbf{aggregation} split.}
\label{fig:error-case-aggregation}
\end{figure*}

\begin{figure*}[t]
\begin{errorcase}[\faExclamationTriangle\;\; Error Case 5 \hfill {\normalfont\ttfamily\footnotesize External Source Type}]

% ── Question ──────────────────────────────────────────────
\noindent{\small\bfseries\textcolor{ec-qcolor}{\faIcon{question-circle}\; Question}}\par\vspace{2pt}
{\small When reviewing the summary of principal suggestions, how many items are designated with the highest urgency classification?}

\ecsep

% ── Gold Answer ───────────────────────────────────────────
\noindent{\small\bfseries\textcolor{ec-gold}{\faIcon{check-circle}\; Gold Answer}}\par\vspace{2pt}
{\small \textbf{4} items (sourced from the Action Items table in \texttt{29\_report} and the Key Feedback Points table in \texttt{44\_report}).}

\ecsep

% ── Retrieved Evidence ────────────────────────────────────
\noindent{\small\bfseries\textcolor{ec-evidence}{\faIcon{search}\; Retrieved Evidence (sample)}}
\begin{enumerate}[leftmargin=1.4em, itemsep=2pt, topsep=3pt, label={\scriptsize[\arabic*]}]
  \item {\small\textit{Attachment (48\_report, 2024-10-30):} Action Items table---``Integrate new case studies into next module'' (\underline{High}); ``Circulate updated EC and academic readings'' (\underline{High}).}
  \item {\small\textit{Attachment (46\_report, 2024-10-28):} ``Expand digital literacy training for teachers''---Priority Level: \underline{High}. {\footnotesize(chunk cuts off after first row)}}
  \item {\small\textit{Attachment (31\_report, 2024-08-15):} Action Items table header visible, but row content truncated at chunk boundary.}
\end{enumerate}

\ecsep

% ── Model Predictions (all incorrect) ────────────────────
\noindent{\small\bfseries\textcolor{ec-titlebg}{\faIcon{robot}\; Model Predictions}}
\vspace{3pt}

\begin{tcolorbox}[enhanced, colback=ec-lightgray, colframe=ec-sepcolor, boxrule=0.3pt, arc=1.5pt,
  left=7pt, right=7pt, top=4pt, bottom=4pt]
{\small\xmark\;\textbf{RAG:}\; At least \textbf{three} items are designated with the highest urgency: (1)~``Integrate new case studies into next module''---High; (2)~``Circulate updated EC and academic readings''---High; (3)~``Expand digital literacy training for teachers''---High.}

\vspace{4pt}
{\small\xmark\;\textbf{Mem0:}\; Based on the available information, it is not possible to determine the number of items classified with the highest urgency.}

\vspace{4pt}
{\small\xmark\;\textbf{Full-context:}\; \textbf{Three} items are designated with the highest urgency classification, based on the Action Items table where ``Urgent hotfix'' and other critical remediation steps are prioritized.}
\end{tcolorbox}

\ecsep

% ── Failure Analysis ──────────────────────────────────────
\noindent{\small\bfseries\textcolor{ec-analysis}{\faIcon{lightbulb}\; Failure Analysis}}\par\vspace{2pt}
{\small The correct count of four requires \emph{cross-document aggregation}: ``High''-priority items are spread across two separate attachment tables (\texttt{29\_report} and \texttt{44\_report}). RAG retrieves partial rows from each {\scriptsize[1--3]} but chunk-boundary truncation hides the fourth item, leading to an undercount of three. Mem0, lacking structured attachment access, abstains entirely. The full-context model also produces three but fabricates unsupported details (``Urgent hotfix,'' ``rapid patching''), hallucinating a justification from unrelated security-themed text rather than the actual education-policy tables.}

\end{errorcase}
\caption{Representative error case from the \textbf{external source} split.}
\label{fig:error-case-attachment}
\end{figure*}

\begin{figure*}[h]
\centering
\begin{prompt}{Prompt for Plan Generation}
Today is \pvar{\{TIMESTAMP\}}. You will assume the role of the character described below and create a comprehensive, realistic plan list for future events from a first-person perspective.

\phead{Character Background}
\ptag{<Background>} \\
\pvar{\{BACKGROUND\}} \\
\ptag{</Background>}

\phead{Existing Commitments}
The following plans are already scheduled. You MUST consider the important existing events in your final plan list. Do not ignore or delete them. If a new plan falls on the same date as an existing plan, you MUST COMBINE them into a single comprehensive plan entry.

\ptag{<ExistingPlans>} \\
\pvar{\{EXIST\}} \\
\ptag{</ExistingPlans>}

\phead{Planning Guidelines}
\textbf{1. Short-Term Plans (Within Next Few Weeks)} \\
Generate 2-4 in total daily activities that reflect the character's routine and personality:
\begin{itemize}[leftmargin=1.5em, noitemsep, topsep=0pt, label=-]
    \item Daily Essentials: Work tasks, meals, commute, exercise, self-care
    \item Social Activities: Meeting friends/family, calls, gatherings
    \item Personal Interests: Hobbies, entertainment, media, shopping
    \item Responsibilities: Errands, appointments, household tasks
\end{itemize}

\vspace{0.3em}
\textbf{2. Long-Term Plans (Within Next Few Months)} \\
Create 1-3 in total significant life events or milestones that align with the character's trajectory:
\begin{itemize}[leftmargin=1.5em, noitemsep, topsep=0pt, label=-]
    \item Career \& Education: Promotions, job changes, graduations
    \item Life Milestones: Marriage, birthdays, relocations
    \item Personal Development: Study abroad, skill acquisition
    \item Health \& Unexpected: Medical procedures, recovery, accidents
\end{itemize}

\textbf{Mandatory Requirements}
\begin{itemize}[leftmargin=1.5em, noitemsep, topsep=0pt]
    \item Date Format: Use DD/MM/YYYY format exclusively.
    \item Date Uniqueness: Each date MUST appear exactly ONCE.
    \item Activity Consolidation: Multiple activities on the same date must be combined.
    \item JSON Structure: Output must be a valid JSON array.
\end{itemize}

\phead{Example Output format}
[\\
\hspace*{1em}\{\{ \\
\hspace*{2em}"Plan": "Morning gym session at 6 AM...", \\
\hspace*{2em}"Date": "07/05/2024", \\
\hspace*{2em}"Category": "Short-Term" \\
\hspace*{1em}\}\}, \\
\hspace*{1em}\{\{ \\
\hspace*{2em}"Plan": "Submit PhD application...", \\
\hspace*{2em}"Date": "02/09/2024", \\
\hspace*{2em}"Category": "Long-Term" \\
\hspace*{1em}\}\} \\
]

Directly present your complete plan as a JSON array without other symbols, ensuring each entry is properly formatted and categorized.
\end{prompt}
\caption{\textbf{Prompt for Plan Generation Module.} This prompt instructs the model to generate a comprehensive plan list for a character based on their background and existing commitments, including short-term and long-term plans.}
\label{fig:prompt-plan-generation}
\end{figure*}
\begin{figure*}
\centering
\begin{prompt}{Profile Update System Prompt}
You are tasked with analyzing an event outcome and the passage of time to identify all necessary factual updates to the user's profile.

Event Outcome: \\
\ptag{<Outcome>} \\
\pvar{\{OUTCOME\}} \\
\ptag{</Outcome>}

Current Profile below include 3 top-level keys: Identity Information, Relationships, Belongings. \\
\ptag{<Profile>} \\
\pvar{\{PROFILE\}} \\
\ptag{</Profile>}

\phead{Update Guidelines}
Updates may include the following aspects:

\vspace{0.3em}
\textbf{1. Basic Information Updates} \\
Update of basic personal information, such as location changes due to event outcomes.

\vspace{0.3em}
\textbf{2. Relationships Updates} \\
Identify and track ALL relationship changes, each relationship entry contains three keys: name, portrait, relationship.
\begin{itemize}[leftmargin=1.5em, noitemsep, topsep=0pt, label=-]
    \item New People: ANY person mentioned in outcome who interacts with the user but is NOT in current relationships
    \item Updated Relationships: Changes in relationship status, closeness, or dynamics with existing contacts
    \item Removed Relationships: People no longer in contact or relationships that have ended
\end{itemize}

\vspace{0.3em}
\textbf{3. Belongings Updates} \\
Identify and track proper belongings changes:
\begin{itemize}[leftmargin=1.5em, noitemsep, topsep=0pt, label=-]
    \item New Items: New items purchased, received, found, or obtained
    \item Updated Items: Changes in item status (repaired, upgraded, expiry, condition changes)
    \item Removed Items: Items sold, lost, given away, broken, unusable or discarded
    \item Categories: Vehicles, computer, phone, book, pet, art, antiques, clothing, jewelry, shoes, bags, sports gear, musical instruments, hand tools, health-related devices, photography equipment. Do NOT invent new categories.
\end{itemize}

\textbf{Analysis Process}
\begin{enumerate}[leftmargin=1.5em, noitemsep, topsep=0pt, label=\arabic*.]
    \item Review Event Outcome: Extract all people, possessions, and factual changes described in the outcome.
    \item Review Profile for Changes: Check for any attributes that should be updated due to time progression or event outcomes.
    \item Determine Required Operations: For each change, specify whether to add, update, or remove item.
    \item Generate Update Function: Create a Python function that implements all necessary changes.
    \item Check function correctness: The function can modify existing values or add/delete entries from existing values, but MUST NOT add new top-level keys or change the JSON structure. Avoid replacing the entire attribute with '='. Do not add function comments.
\end{enumerate}

Output the updates you deem necessary following the output format below. Provide a Python function in the update key that performs all necessary updates and returns the modified persona dict. If no updates are needed, directly return the persona.

\phead{Output format}
\{\{ \\
\hspace*{1.5em}"update": "def update\_persona(persona):\\    return persona", \\
\hspace*{1.5em}"reason": "Explanation of the reason..." \\
\}\}
\end{prompt}
\caption{\textbf{Prompt for Profile Update Module.} This module handles factual attribute changes (e.g., relationship, beloinginds) and enforces strict JSON schema constraints.}
\label{fig:prompt-update}
\end{figure*}
\begin{figure*}[h]
\centering
\begin{prompt}{Traits \& Status Update System Prompt}
You are tasked with analyzing an event outcome to determine necessary updates to the user's traits, recent status attributes.

Event Outcome: \\
\ptag{<Outcome>} \\
\pvar{\{OUTCOME\}} \\
\ptag{</Outcome>}

Current Profile below include 4 categories: Identity Information, Traits, Personality, Current Status. \\
\ptag{<Profile>} \\
\pvar{\{PROFILE\}} \\
\ptag{</Profile>}

\phead{Update Guidelines}
The words enclosed in \texttt{**} below are all top-level keys in the JSON profile.

\vspace{0.3em}
\textbf{1. \texttt{Traits} Updates} \\
Identify key changes in personal characteristics:
\begin{itemize}[leftmargin=1.5em, noitemsep, topsep=0pt, label=-]
    \item \texttt{hobbies} (dictionary list): New activities discovered, abandoned hobbies, or modified hobby descriptions
    \item \texttt{personal\_preferences} (dictionary list): Changes in preference levels (Scale: "strongly dislike" to "strongly like")
    \item \texttt{lifestyle} (dictionary): Significant modifications to daily routines or living patterns
\end{itemize}

\vspace{0.3em}
\textbf{2. \texttt{Current Status} Updates} \\
Track the character's current state and recent significant developments:
\begin{itemize}[leftmargin=1.5em, noitemsep, topsep=0pt, label=-]
    \item \texttt{health\_status} (string): Update current physical and mental health condition
    \item \texttt{mood} (string): Update current emotional state if significantly affected
    \item \texttt{ongoing\_events} (dictionary list): Significant events currently ongoing (work deadlines, trips). Outdated events need to be deleted.
\end{itemize}

\phead{STRICT KEY REQUIREMENT}
\begin{itemize}[leftmargin=1.5em, noitemsep, topsep=0pt, label=-]
    \item \texttt{ongoing\_events}: Each entry MUST contain EXACTLY these 3 keys:
    \begin{itemize}[leftmargin=1.5em, noitemsep]
        \item \texttt{"event"}: event description (string)
        \item \texttt{"started"}: start date in "YYYY-MM-DD" format
        \item \texttt{"expected\_end"}: end date in "YYYY-MM-DD" format
    \end{itemize}

    \item \texttt{hobbies}: List of dictionaries containing:
    \begin{itemize}[leftmargin=1.5em, noitemsep]
        \item \texttt{"origin"}: how/when started (string)
        \item \texttt{"frequency"}: how often practiced (string)
        \item \texttt{"skill\_level"}: proficiency level (string)
    \end{itemize}

    \item \texttt{personal\_preferences}: List of dictionaries with one key-value pair.

    \item \texttt{lifestyle}: Dictionary with keys \texttt{"sleep\_pattern"}, \texttt{"diet"}, \texttt{"exercise"}, \texttt{"social"}, \texttt{"digital\_habits"}.
\end{itemize}

\phead{Analysis Process}
\begin{enumerate}[leftmargin=1.5em, noitemsep, topsep=0pt, label=\arabic*.]
    \item Extract Changes from Outcome: Identify all traits shifts and status changes.
    \item Compare with Current Profile: Review profile to identify attributes needing updates.
    \item Determine Update Operations: Specify add, modify, or remove operations.
    \item Generate Update Function: Create a Python function implementing changes.
    \item Check function correctness: MUST NOT add new top-level keys or change JSON structure. Avoid replacing entire attribute with '='.
\end{enumerate}

Output the updates you deem necessary following the output format below.

\phead{Output format}
\{\{ \\
\hspace*{1.5em}"update": "def update\_persona(persona):\\n    return persona", \\
\hspace*{1.5em}"reason": "Explanation of the reason..." \\
\}\}
\end{prompt}
\caption{\textbf{Prompt for Traits \& Status Update.} This module handles abstract attribute changes (e.g., mood, lifestyle) and enforces strict JSON schema constraints.}
\label{fig:prompt-traits}
\end{figure*}
\begin{figure*}[h]
\centering
\begin{prompt}{Attachment Generation System Prompt}
You are an expert document analyst tasked with identifying realistic digital attachments that would naturally be associated with a specific event outcome of the user. Your goal is to generate metadata for documents that can later be created by a document generation model.

First, carefully analyze the event outcome from the user (using third-person perspective): \\
\ptag{<Outcome>} \\
\pvar{\{OUTCOME\}} \\
\ptag{</Outcome>}

Based on the outcome, identify the most significant details and specify realistic digital attachments that would logically be associated with this event. Focus ONLY on the most relevant documents that would naturally exist.

Format your response as a JSON object with the following categories:
\begin{itemize}[leftmargin=1.5em, noitemsep, topsep=0pt, label=-]
    \item \texttt{"emails"}: Email communications related to the event.
    \item \texttt{"reports"}: Work and study-related documents (e.g., project reports, meeting minutes).
    \item \texttt{"notes"}: Personal daily life notes (e.g., itineraries, shopping lists).
\end{itemize}

\phead{Use this exact JSON structure}
\{\{ \\
\hspace*{1.5em}"emails": [ \\
\hspace*{3.0em}\{\{ \\
\hspace*{4.5em}"sender": "Full Name <email@domain.com>", \\
\hspace*{4.5em}"recipient": "Full Name <email@domain.com>", \\
\hspace*{4.5em}"subject": "Descriptive email subject line", \\
\hspace*{4.5em}"summary": "Detailed summary of email content..." \\
\hspace*{3.0em}\}\} \\
\hspace*{1.5em}], \\
\hspace*{1.5em}"reports": [ \\
\hspace*{3.0em}\{\{ \\
\hspace*{4.5em}"type": "Specific document type...", \\
\hspace*{4.5em}"utterance": "Detailed generation instruction..." \\
\hspace*{3.0em}\}\} \\
\hspace*{1.5em}], \\
\hspace*{1.5em}"notes": [ \\
\hspace*{3.0em}\{\{ \\
\hspace*{4.5em}"type": "Specific document type...", \\
\hspace*{4.5em}"utterance": "Detailed generation instruction..." \\
\hspace*{3.0em}\}\} \\
\hspace*{1.5em}] \\
\}\}

\phead{Document Generation Guidelines}
\begin{enumerate}[leftmargin=1.5em, noitemsep, topsep=0pt, label=\arabic*.]
    \item Analyze the event outcome thoroughly to determine which attachment types are logically relevant. For irrelevant categories, provide empty arrays. For "reports" and "notes", provide at most 2 items each.
    \item Ensure that there are no duplications or inconsistencies between attachments.
    \item Each "utterance" field should contain comprehensive instructions including format requirements, real details, and special formatting. Use professional, clear statements throughout in the first-person user perspective.
    \item Ensure all attachments are contextually appropriate and would realistically exist.
    \item Make attachments practically useful for the character's situation.
\end{enumerate}

Provide your response in JSON format within the Output tags: \\
\ptag{<Output>} \\
\hspace*{1em}[Generate the detailed attachment metadata JSON here] \\
\ptag{</Output>}
\end{prompt}
\caption{\textbf{Prompt for Attachment Utterance Generation.} The model identifies logical digital artifacts (emails, attachments) implied by the event outcome and generates generation utterances for them.}
\label{fig:prompt-attachment}
\end{figure*}
\begin{figure*}[h]
\centering
\begin{prompt}{Standard Query Response Prompt}
You are an AI assistant that helps analyze user history. Based on the relevant evidence below, please provide the answer to the user's query.

Relevant Evidence: \\
\pvar{\{CONTEXT\}} \\

Question Date: \pvar{\{DATE\}} \\
User Query: \pvar{\{QUERY\}}

\vspace{0.5em}
Please provide a concise and accurate answer.

\vspace{0.5em}
Answer:
\end{prompt}
\caption{\textbf{Prompt for Standard Questions.} This template handles general inquiries based on the retrieved user history context.}
\label{fig:prompt-standard}
\end{figure*}
\begin{figure*}[htbp]
\centering
\begin{prompt}{Elaborative Query Response Prompt}
You are an AI assistant that helps analyze user history. Based on the relevant evidence below, please provide the answer to the user's query.
Relevant Evidence: \\
\pvar{\{CONTEXT\}} \\

Question Date: \pvar{\{DATE\}} \\
User Query: \pvar{\{QUERY\}} \\

\begin{enumerate}[leftmargin=1.5em, noitemsep, topsep=0.5em, label=\arabic*.]
    \item If the question contains factual errors, false premises, or contradicts the user's state, explicitly point out the error and propose a compliant alternative.
    \item If the question cannot be answered based on the evidence, state that clearly.
    \item Otherwise, answer the question directly based on the evidence.
\end{enumerate}

Answer:
\end{prompt}
\caption{\textbf{Prompt for Elaborative Real-world Questions.} This template is used for real-world contextual reasoning questions like \textit{Hallucination} and \textit{Misleading} types.}
\label{fig:prompt-adversarial}
\end{figure*}
\begin{figure*}[htbp]
\centering
\begin{prompt}{LLM-as-Judge Evaluation Prompt}
You are an expert evaluator assessing an AI assistant's answer against a reference answer.

Question: \pvar{\{QUESTION\}} \\
Expected Answer: \pvar{\{ANSWER\}} \\
AI Assistant's Answer: \pvar{\{PREDICT\}} \\

Please evaluate the AI assistant's answer on the following criteria: \\

\textbf{\#\#\# Accuracy (0 or 1)}
\begin{itemize}[leftmargin=1.5em, noitemsep, topsep=0.2em]
    \item \textbf{1 (Correct)}: The assistant's answer conveys the same key factual information as the reference answer. Minor differences in wording, additional context, or extra details are acceptable as long as the core facts match.
    \item \textbf{0 (Incorrect)}: The assistant's answer contradicts the reference answer, omits critical facts, provides wrong information, or fails to answer the question. If the question asks about a false premise and the reference answer points out the error, the assistant must also identify the error to be considered correct.
\end{itemize}

\vspace{0.5em}
\textbf{\#\#\# Overall Quality (0-10)}
\begin{itemize}[leftmargin=1.5em, noitemsep, topsep=0.2em]
    \item \textbf{9-10}: Accurate, complete, well-structured, and directly addresses the question.
    \item \textbf{7-8}: Accurate with minor omissions or slightly verbose.
    \item \textbf{5-6}: Partially correct; contains some relevant information but misses key points.
    \item \textbf{3-4}: Mostly incorrect but shows some understanding of the topic.
    \item \textbf{1-2}: Almost entirely wrong or irrelevant.
    \item \textbf{0}: Completely wrong, refuses to answer, or is incoherent.
\end{itemize}

\vspace{0.5em}
Provide your evaluation in the following JSON format: \\
\{\{ \\
\hspace*{2em} "accuracy\_score": [0 or 1], \\
\hspace*{2em} "overall\_score": [0-10] \\
\}\} \\

Only respond with valid JSON, no other text.
\end{prompt}
\caption{\textbf{Prompt for LLM-as-Judge Evaluation.} This template instructs the evaluator model to assess the accuracy and quality of the generated response against a ground truth reference. Note we use accuracy as a strict binary metric for better evaluation.}
\label{fig:prompt-llm-judge}
\end{figure*}
\begin{figure*}[t]
\centering
\begin{prompt}{Bullet Point Classification}
You are an expert in conversation design. Your task is to classify bullet points from a user's day into appropriate message types for a realistic dialogue with their personal assistant. \\
The goal is to simulate a natural and varied interaction. Therefore, deeply analyze the user's likely intent and emotional state for each event.

\phead{\#\# Available Message Types}
\begin{itemize}[leftmargin=1.5em, noitemsep, topsep=0pt, label=$\bullet$]
    \item \textbf{information\_sharing}: For sharing what happened, reporting outcomes, giving updates about events
    \item \textbf{advice\_seeking}: For asking for advice, seeking new ideas, requesting guidance about decisions or problems
    \item \textbf{status\_update}: For reflecting on experiences, sharing feelings, explaining how perspectives or preferences have changed
    \item \textbf{scheduling}: For discussing future plans, coordinating upcoming activities, or planning next steps
\end{itemize}

\phead{\#\# Bullet Points to Classify}
\pvar{\{BULLETS\}}

\phead{\#\# Instructions}
For each bullet point, think about the most appropriate message type if the user were to chat with their assistant. \\
Analyze the content to decide:
- The natural flow of conversation and the event context
- Is the user seeking help, sharing information, expressing feelings, or planning the future?
- How does this point fit into the user's daily life and the emotional context?

\vspace{0.5em}
Please ensure a reasonable and natural distribution of these types based on the content.

\phead{\#\# Output Format}
Please respond in the following JSON format: \\
\ptag{[} \\
\quad \ptag{\{} \\
\quad\quad "bullet\_point": "exact bullet point text", \\
\quad\quad "message\_type": "one of: information\_sharing, advice\_seeking, status\_update, scheduling", \\
\quad\quad "reason": "brief explanation including the likely timing and intent" \\
\quad \ptag{\}} \\
\ptag{]}
\end{prompt}
\caption{\textbf{Prompt for Bullet Point Classification.} This prompt classifies daily event bullet points into dialogue categories to guide the generation of diverse, intent-aware conversations between the user and their personal assistant. The attachment consultation type is directly inserted into each day's conversation.}
\label{fig:prompt-bullet-classify}
\end{figure*}
\begin{figure*}[h]
\centering
\begin{prompt}{User Simulation: Conversation Initiator}
You need to play a role as a real user in a conversation with your personal AI assistant. Start sharing about specific topic with what happened today, this is the beginning of a conversation about this topic.

\phead{\#\# Background}
User profile and communicate style: \pvar{\{PROFILE\}} \\
Current time: \pvar{\{TIME\}} \\
What happened during the whole day: \\
\pvar{\{OUTCOME\}}

Conversation History: \\
--- \\
\pvar{\{HISTORY\}} \\
---

Current conversation topic: \\
\pvar{\{TOPIC\}}

\phead{\#\# Guidelines}
\begin{itemize}[leftmargin=1.5em, noitemsep, topsep=0pt, label=-]
    \item Introduce the situation naturally, as if you're settling in to tell the full story about the topic
    \item Don't rush to share everything at once - leave room for follow-up
    \item Use your natural speaking style based on your personality
    \item Do NOT repeat factual details or events that have already been explicitly shared in the "Conversation History" above.
    \item Try to include granular details including time, person, event, location: such as "this morning", a colleague's name, etc.
\end{itemize}

\vspace{0.5em}
\texttt{IMPORTANT}: FOCUS on and talk about the current topic only. Don't switch to a different topic or activity happening today.
\end{prompt}
\caption{\textbf{Prompt for User Simulator (Initial Turn).} This prompt initializes the conversation based on the daily event summary and communication-related user profile attributes (Prompts may vary slightly across different dialogue categories, the overall template remains consistent).}
\label{fig:prompt-user-init}
\end{figure*}
\begin{figure*}[h]
\centering
\begin{prompt}{User Simulation: Follow-up Turn}
You are roleplaying as a real user in an ongoing conversation with your personal AI assistant.

\phead{\#\# Background}
User profile and communication style: \pvar{\{PROFILE\}} \\
Current time: \pvar{\{TIME\}}

Conversation History: \\
--- \\
\pvar{\{HISTORY\}} \\
---

CURRENT TOPIC: \\
\pvar{\{TOPIC\}}

\phead{\#\# Your Goal}
Continue the conversation naturally based \textbf{ONLY} on the information provided in the CURRENT TOPIC and Conversation History.

\textbf{Critical Anti-Hallucination Rules}
\begin{enumerate}[leftmargin=1.5em, noitemsep, topsep=0pt, label=\arabic*.]
    \item NO NEW FACTS: You are STRICTLY FORBIDDEN from inventing, assuming, or adding any factual details (events, names, places, times, objects) that are not explicitly stated in the "CURRENT TOPIC" text above.
    \item Deepen, Don't Expand: Instead of adding new plot points, deepen the conversation by:
    \begin{itemize}[leftmargin=1em, label=$\cdot$]
        \item Expressing your feelings or opinions about the facts already shared.
        \item Asking the assistant for advice or analysis based on the known facts.
        \item Clarifying or re-emphasizing a detail you already mentioned if the assistant misunderstood.
    \end{itemize}
    \item Stay within Boundaries: If you have shared all the factual details from the CURRENT TOPIC, do NOT make up more. Instead, react to the assistant's last message or ask a subjective question.
\end{enumerate}

\phead{Your Follow-up Options}
Choose ONE approach:
\begin{itemize}[leftmargin=1.5em, noitemsep, topsep=0pt]
    \item Share Remaining Details: If there are specific facts in the "CURRENT TOPIC" you haven't mentioned yet, share them now.
    \item React \& Reflect: If all facts are shared, react to the assistant's advice or comment using your persona's tone (e.g., agree, disagree, express relief/worry).
    \item Ask for Input: Ask a practical question based only on the details already established (e.g., "Given [known fact], what should I do?").
\end{itemize}

\textbf{Response Requirements}
\begin{itemize}[leftmargin=1.5em, noitemsep, topsep=0pt, label=-]
    \item Concise: Keep your response focused and to the point.
    \item Natural Tone: Casual, direct, matching your User Profile. No formal pleasantries.
    \item Decisive: State your thought clearly. Do not ask "what do you think?" unless you are genuinely asking for specific advice on a known problem.
\end{itemize}
\end{prompt}
\caption{\textbf{Prompt for User Simulator (Follow-up Turn).} This prompt further deepens the user-assistant conversational scenarios.}
\label{fig:prompt-user-followup}
\end{figure*}
\begin{figure*}[htbp]
\centering
\begin{prompt}{Challenging QA Generation Prompt}
You are an expert in question answering. Your task is to assume the role of \pvar{\{CHARACTER\}} and generate challenging QA pairs in first-person perspective from \pvar{\{DATE\}} that effectively assess a chat assistant's long-term memory capabilities using the provided evidence and question type definition.

\phead{Input Context}
\textbf{1. Question Type Definition:} \\
\pvar{\{QUESTION DEFINITION\}}

\textbf{2. Relevant Evidence:} \\
\pvar{\{EVIDENCE\}}

\phead{Guidelines}
Adhere to the following requirements to create the QA pairs:

\textbf{\#\#\# 1. Question Requirements}
\begin{itemize}[leftmargin=1.5em, noitemsep, topsep=0.2em, label=-]
    \item \textbf{Consistency}: Strictly follow the definition and challenging characteristics provided above. Contextualize temporally as of \pvar{\{DATE\}}.
    \item \textbf{Perspective}: Use natural FIRST-PERSON tone (e.g., "What did I do...").
    \item \textbf{Single-Target Focus}: Ask for ONE specific piece of information (fact, name, date, etc.).
    \item \textbf{Avoid Compound Questions}: DO NOT use "and" to combine questions.
    \item \textbf{Avoid Open-Ended Analysis}: No "How did this affect me?" or "Why did I do this?".
\end{itemize}

\textbf{\#\#\# 2. Answer Requirements}
\begin{itemize}[leftmargin=1.5em, noitemsep, topsep=0.2em, label=-]
    \item \textbf{Conciseness}: Provide short answers supported by evidence (phrase or few words).
    \item \textbf{Accuracy}: Must be strictly supported by the provided evidence.
\end{itemize}

\textbf{\#\#\# 3. Difficulty Requirements} \\
Generate questions where the \textit{path to the answer} is hard, but the \textit{answer itself} is concise. You may consider:
\begin{itemize}[leftmargin=1.5em, noitemsep, topsep=0.2em, label=-]
    \item Synthesize evidence from different contexts or distant time periods.
    \item Require multi-hop inference chains or logical deduction.
    \item Ensure answers cannot be directly extracted without reasoning.
\end{itemize}

\textbf{\#\#\# 4. Characteristics Requirements}
\begin{itemize}[leftmargin=1.5em, noitemsep, topsep=0.2em, label=-]
    \item \textbf{Source}: Select ONLY from the characteristics defined in the Question Type Definition.
    \item \textbf{Specificity}: Use the exact terminology provided in the Question Type Definition.
\end{itemize}

\phead{Output Format}
Generate exactly 3 QA pairs using this JSON structure: \\
{[} \\
\hspace*{1em} \{\{ \\
\hspace*{2em} "question": "Natural first-person question you would ask", \\
\hspace*{2em} "question\_date": "YYYY-MM-DD", \\
\hspace*{2em} "answer": "Concise, specific answer", \\
\hspace*{2em} "characteristics": ["List of specific challenging characteristics"], \\
\hspace*{1em} \}\} \\
{]}
\end{prompt}
\caption{\textbf{Prompt for Challenging QA Generation.} This prompt instructs the model to generate high-difficulty QA pairs. The challenging characteristics definitions are added in the question definition input.}
\label{fig:prompt-qa-gen}
\end{figure*}

\end{document}